\pdfoutput=1

\documentclass[11pt]{article}

\usepackage[]{ACL2023}

\usepackage{times}
\usepackage{latexsym}

\usepackage[T1]{fontenc}

\usepackage[utf8]{inputenc}

\usepackage{microtype}

\usepackage{inconsolata}

\usepackage{graphicx}
\usepackage{amsmath}
\usepackage{amssymb}
\usepackage{booktabs}

\usepackage{color}
\usepackage[caption=false]{subfig}
\usepackage{amsfonts}
\usepackage{graphicx}

\usepackage{multirow,diagbox,makecell}
\usepackage{tikz}

\usepackage{pifont}
\newcommand{\cmark}{\ding{51}}%
\newcommand{\xmark}{\ding{55}}%
\usepackage{booktabs}

\usepackage{capt-of}

\title{Aerial Vision-and-Dialog Navigation}

\author{Yue Fan, Winson Chen, Tongzhou Jiang, Chun Zhou, Yi Zhang, Xin Eric Wang \\University of California, Santa Cruz \\ \{yfan71, wchen157, tojiang, czhou43, yiz, xwang366\}@ucsc.edu}

\begin{document}
\maketitle
\begin{abstract}
The ability to converse with humans and follow natural language commands is crucial for intelligent unmanned aerial vehicles (a.k.a. drones). 
It can relieve people's burden of holding a controller all the time, allow multitasking, and make drone control more accessible for people with disabilities or with their hands occupied. To this end, we introduce Aerial Vision-and-Dialog Navigation (AVDN), to navigate a drone via natural language conversation. 
We build a drone simulator with a continuous photorealistic environment and collect a new AVDN dataset of over 3k recorded navigation trajectories with asynchronous human-human dialogs between commanders and followers. 
The commander provides initial navigation instruction and further guidance by request, while the follower navigates the drone in the simulator and asks questions when needed. 
During data collection, followers' attention on the drone's visual observation is also recorded. 
Based on the AVDN dataset, we study the tasks of aerial navigation from (full) dialog history and propose an effective Human Attention Aided Transformer model (HAA-Transformer), which learns to predict both navigation waypoints and human attention. 
Dataset and code are released: \url{https://sites.google.com/view/aerial-vision-and-dialog/home}.
\end{abstract}

\section{Introduction}


\begin{figure}[t]

        \centering
        \includegraphics[width=1\linewidth]{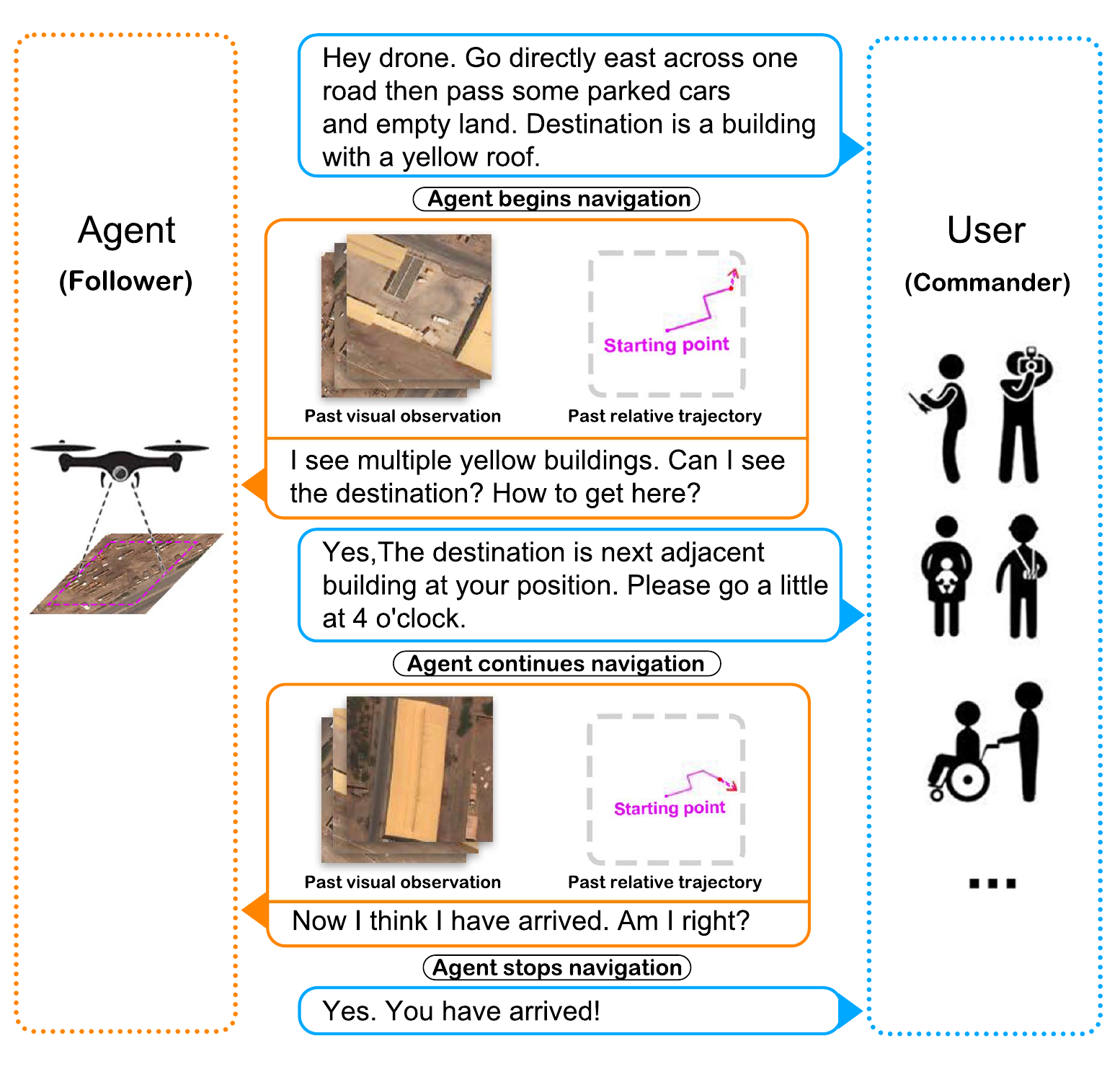}
        \setlength{\abovecaptionskip}{-0.5cm}
        \setlength\tabcolsep{0pt}
        \caption{
        An example of Aerial Vision-and-Dialog Navigation (AVDN). 
        The user instructs the agent to fly to a destination. During the navigation, the agent can ask questions while showing the images of past visual observations and relative trajectory. The user will talk back at a convenient time to provide further guidance to the agent without having to monitor the agent all the time. 
        }
        \label{fig:intro_fig}

\end{figure}

Drones have been widely adopted for many applications in our daily life, from personal entertainment to professional use. It has the advantage of mobility and observing large areas over ground robots. However, compared with ground robots, the control of the aerial robot is more complex because an extra degree of freedom, altitude, is involved. To control a drone, people often need to hold a controller all the time, so it is essential to create a hands-free control experience for drone users and develop an intelligent drone that can complete tasks simply by talking to humans. It can lower the barrier of drone control for users with some disabilities and who have their hands occupied by activities such as taking photos, writing, etc.

Therefore, this work introduces Aerial Vision-and-Dialog Navigation (AVDN), aiming to develop an intelligent drone that can converse with its user to fly to the expected destination. 
As shown in Figure \ref{fig:intro_fig}, the user (commander) provides instructions, and the aerial agent (follower) follows the instruction and asks questions when needed. The past visual trajectories are also provided along with the question, which frees the commander from monitoring the drone all the time and minimizes the burden of drone control. 
In this free-form dialog, potential ambiguities in the instruction can be gradually resolved through the further instructions provided by the commanders upon request. 

To implement and evaluate the AVDN task, we build a photorealistic simulator with continuous state space to simulate a drone flying with its onboard camera pointing straight downward. Then we collect an AVDN dataset of 3,064 aerial navigation trajectories with human-human dialogs, where crowd-sourcing workers play the commander role and drone experts play the follower role, as illustrated in Figure~\ref{fig:intro_fig}. Moreover, we also collect the attention of human followers over the aerial scenes for a better understanding of where humans ground navigation instructions. 

Based on our AVDN dataset, we introduce two challenging navigation tasks, Aerial Navigation from Dialog History (ANDH) and Aerial Navigation from Full Dialog History (ANDH-Full). Both tasks focus on predicting navigation actions that can lead the agent to the destination area, whereas the difference is that ANDH-Full presents the agent with full dialog and requires it to reach the final destination~\cite{kim2021ndh}, while ANDH evaluates the agent's completion of the sub-trajectory within a dialog round given the previous dialog information~\cite{thomason2020vision}.


The proposed tasks open new challenges of sequential action prediction in a large continuous space and natural language grounding on photorealistic aerial scenes. 
We propose a sequence-to-sequence Human Attention Aided Transformer model (HAA-Transformer) for both tasks. 
The HAA-Transformer model predicts waypoints to reduce the complexity of the search space and learns to stop at the desired location. More importantly, it is jointly trained to predict human attention from the input dialog and visual observations and learns where to look during inference.  
Experiments on our AVDN dataset show that multitask learning is beneficial and human attention prediction improves navigation performance.
The main contributions are concluded as follows:
\begin{itemize}
    \item We create a new dataset and simulator for aerial vision-and-dialog navigation. The dataset includes over 3K aerial navigation trajectories with human-human dialogs.  
    \item  We introduce ANDH and ANDH-Full tasks to evaluate the agent's ability to understand natural language dialog, reason about aerial scenes, and navigate to the target location in a continuous photorealistic aerial environment.
    \item  We propose an HAA-Transformer model as the baseline for ANDH and ANDH-Full. Besides predicting the waypoint navigation actions, HAA-Transformer also learns to predict the attention of the human follower along the navigation trajectory. Experiments on our AVDN dataset validate the effectiveness of the HAA-Transformer model.
\end{itemize}

\section{Related work}



\paragraph{Vision-and-Language Navigation} 
Vision-and-Language Navigation (VLN) is an emerging multimodal task that studies the problem of using both language instructions and visual observation to predict navigation actions. We compare some of the works with our AVDN dataset in Table \ref{compare_datasets}. 
Early VLN datasets such as \citet{anderson2018vision,ku2020room} start with the indoor house environments in the Matterport3D simulator \cite{chang2017matterport3d}, where the visual scenes are connected on a navigation graph. To simulate continuous state change as in the real world, \citet{krantz2020beyond} built a 3D continuous environment by reconstructing the scene based on topological connections where the agent uses continuous actions during the navigation. Some other VLN studies focus on language instructions. \citet{nguyen2019vision,nguyen2019help,thomason2020vision} created datasets where the agent can interact with the user by sending fixed signals or having dialogs.
There are also works on synthetic indoor environments, such as \citet{shridhar2020alfworld,padmakumar2021teach} that use an interactive simulation environment with synthetic views named ALFRED, where the agent needs to follow language instructions or dialogs to finish household tasks. 
Besides the indoor environment, some VLN datasets work on the more complex outdoor environment, such as the Touchdown dataset \cite{chen2019touchdown} and the modified LANI dataset \cite{misra2018mapping}. \citet{blukis2019learning} is similar to ours for both using drones. However, the synthetic environment used has a gap from the realistic scene, and they ignored the control of the drone's altitude, where such navigation is oversimplified and has a large gap towards navigation in the real world in terms of language and vision aspects. Our work absorbs the advantage from previous works where we have continuous environments and dialog instructions to better approximate the real-world scenario. 

\begin{table}[t]
\setlength{\abovecaptionskip}{0.1cm}
\setlength\tabcolsep{0pt}
    \centering
    {
    \resizebox{\linewidth}{!}{
        \begin{tabular}{lccccc}
        \toprule 
        \textbf{Dataset} 
        & \textbf{Env}
        & \textbf{Photorealistic}
        & $\begin{array}{c}
         \textbf{Continuous} \\
         \textbf{Space}
        \end{array}$
        & \textbf{Dialog}
        & $\begin{array}{c}
         \textbf{Free} \\
         \textbf{Form}
        \end{array}$
        \\
        \midrule 
        R2R, RxR & indoor &\cmark & \xmark & \xmark & \cmark \\
        VNLA, HANNA & indoor& \cmark & \xmark & \cmark & \xmark \\
        VLN-CE & indoor & \xmark & \cmark & \xmark & \cmark   \\
        CVDN & indoor &  \cmark & \xmark & \cmark & \cmark \\
        ALFRED & indoor & \xmark & \xmark & \xmark & \cmark \\
        DialFRED & indoor & \xmark & \xmark & \cmark & \xmark \\
        TEACh& indoor &  \xmark & \cmark & \cmark & \cmark \\
        TouchDown& street-view  & \cmark & \xmark & \xmark & \cmark \\
        modified LANI & aerial & \xmark & \cmark & \xmark & \cmark \\
        \midrule 
        AVDN (ours) & aerial & \cmark & \cmark & \cmark & \cmark \\
        \bottomrule
        \end{tabular}
        }
    }
    \label{tab:datasets}
    \caption{Example Vision-and-Language Navigation Datasets. R2R \cite{anderson2018vision}, RxR \cite{ku2020room}, VNLA \cite{nguyen2019vision}, HANNA \cite{nguyen2019help}, VLN-CE \cite{krantz2020beyond}, CVDN
    \cite{thomason2020vision}, 
    ALFRED \cite{ALFRED20}, DialFRED \cite{gao2022dialfred}, TEACh \cite{padmakumar2021teach}, TouchDown \cite{chen2019touchdown}, modified LANI \cite{blukis2019learning}.
    \label{compare_datasets}}
\end{table}

\paragraph{Aerial Navigation}
Using both vision and language for aerial navigation is a less studied topic, whereas vision-only aerial navigation for drones is already an active topic in the field. Some inspiring works \cite{loquercio2018dronet,giusti2015machine,smolyanskiy2017toward,fan2020learn,Ilker2020multimodal,Amdras2017Zurich,kang2019generalization} worked on using pre-collected real-world drone data to tackle aerial vision navigation problems. Due to the hardness of collecting data and the risk of crashes, some other works applied simulation for aerial navigation \cite{chen2018learning,Shah2017AirSim,chen2020valid}, where rich ground truths are provided without the need for annotation. However, the modality of language is missing in these prior works and as a result, the navigation tasks only contain simple goals. As for the aerial vision-and-language navigation task in this work, the navigation is guided by natural dialog. As a result, it allows more diverse and complex navigation and also resolves ambiguities during complicated navigation.


\section{Dataset}
\begin{figure}[t]
\setlength{\abovecaptionskip}{0.1cm}
    \centering
        \includegraphics[width=0.48\textwidth]{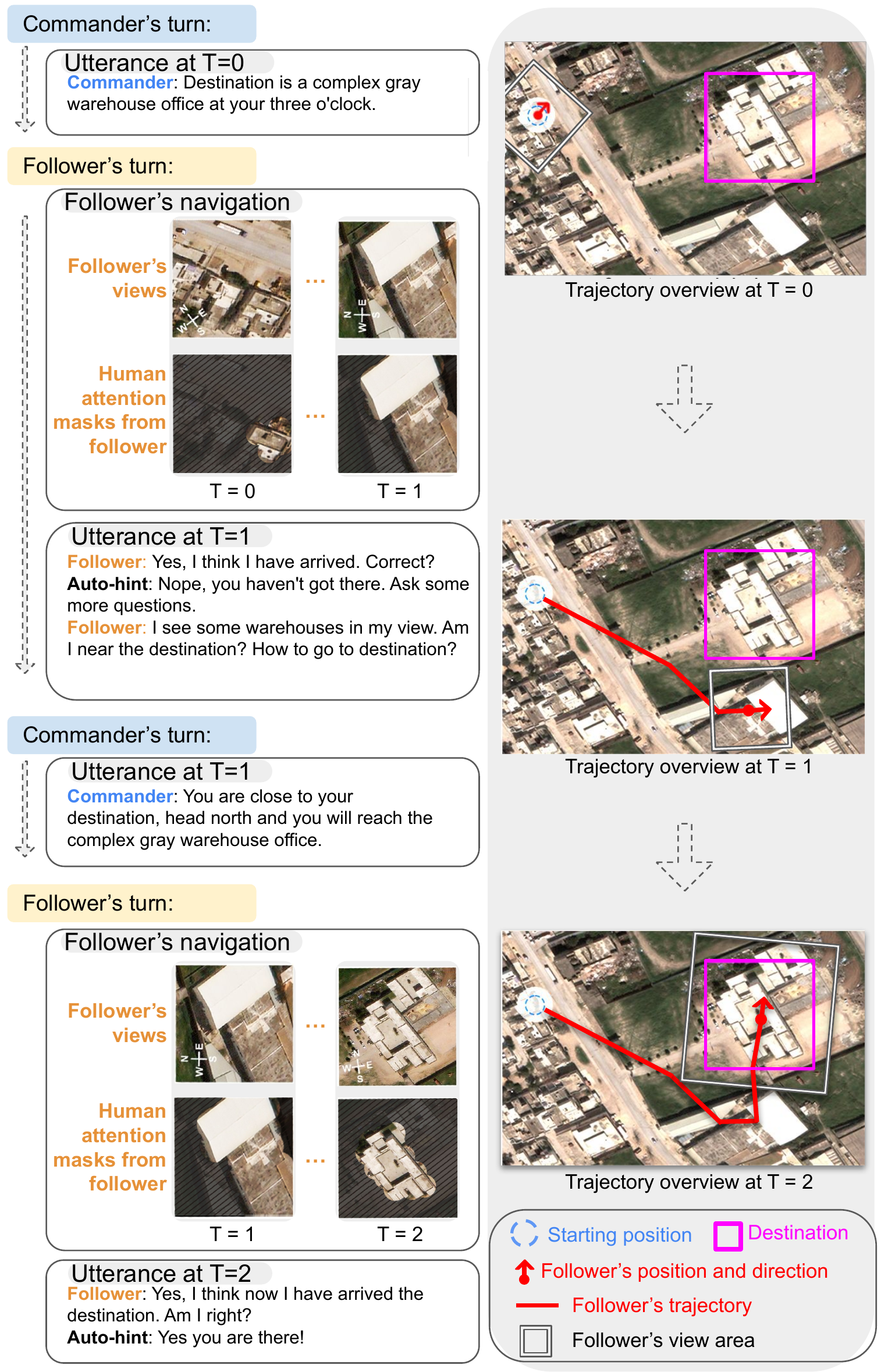}
        \caption{Example of a trajectory in our AVDN dataset. On the left, the commander's turn and the follower's turn alternate in chronological order. In each turn, dialog utterances are shown, and the follower's turn also shows the navigation process that spans from time step $T$ to $T+1$, including the follower's observation and attention. On the right, there are trajectory overviews at different time steps. More examples can be found in the Appendix.}
        \label{fig:figure_2}
\end{figure}

The AVDN dataset includes dialogs, navigation trajectories, and the drone's visual observation with human attention, where an example is shown in Figure \ref{fig:figure_2}. With the help of a newly proposed simulator, we record the AVDN trajectories created by two groups of humans interacting with each other, playing either the commander role or the follower role.
Our AVDN dataset is the first aerial navigation dataset based on dialogs to the best of our knowledge. 

\subsection{Simulator}

We build a simulator to simulate the drone with a top-down view area. Our simulation environment is a continuous space so that the simulated drone can move continuously to any point within the environment. The drone's visual observations are square images generated corresponding to the drone's view area by cropping from high-resolution satellite images in the xView dataset \cite{lam2018xview}, an open-source large-scale satellite image object detection dataset. In this way, our simulator is capable of providing continuous frames with rich visual features. We also design an interface for our simulator, where the simulated drone can be controlled with a keyboard and the drone's visual observation will be displayed in real-time with a digital compass.
During the control, users can also provide their attention over the displayed images on the interface by clicking the region they attend to. Last but not least, our simulator is capable of generating trajectory overviews, i.e. commander's view, showing the starting positions, destination areas, current view area and past trajectory (if exists) as in Figure~\ref{fig:figure_2}.

        


\subsection{Dataset Structure}
In our AVDN dataset, each navigation trajectory includes time steps $T = 0,1, \dots, M$, where $M \geq 1$.
At $T = 0$, an initial instruction is provided by the commander. Between adjacent time steps, there is a corresponding navigation sub-trajectory. At every time step of $0<T < M$, there are questions from the follower and the corresponding answers from the commander. At $T = M$, the navigation trajectory ends because the destination area $Des$ is reached and claimed by the follower. For details about when a trajectory ends, please refer to Section~\ref{success_condition} Success Condition.

There are $M$ follower's view area sequences $<u^T_{0}, u^T_{1}, \dots, u^T_{N_T}>$, $N_T$ is the length of $T$-th sequence, where the view area's center coordinate $c^T_{i}$ always falls on the trajectory. Therefore, based on each view area, we could retrieve not only the simulated drone's location $c_i$, but also direction $d_i$ and altitude $h_i$. Last but not least, for each view area $u$, there is a corresponding binary human attention mask with the same size. The area in $u$ that corresponds to the white area on the mask is where the follower attended.


\subsection{Dataset Collection}
\label{datacollection}
We collect our dataset with the help of Amazon Mechanical Turk (AMT) workers and drone experts, where AMT workers play the commander role to provide instructions and drone experts play the follower role to control a simulated drone and carry out the instruction. We pay the workers with wages no less than \$15/h, and the data collection lasts for 90 days.  
We adopt an asynchronous data collection method, where the followers and commanders work in turns rather than simultaneously. This not only lowers the cost of data collection but also simulates how aerial vision-and-dialog navigation would work in practice, where the commanders will not monitor the follower's actions all the time.  

\paragraph{Pipeline}
Before the start of data collection, we first sample objects in the xView dataset \cite{lam2018xview} as the destination areas and pair them with randomly selected initial follower's view areas within 1.5km distance. Then, using our simulator, we generate the trajectory overview at time step $T = 0$, as shown in Figure \ref{fig:figure_2}, which becomes the initial commander's view.

During data collection, the initial commander's view is presented to AMT workers for creating the initial instructions. We instruct the AMT workers to write instructions as if they are talking to a drone pilot based on the marked satellite images. 
Next, we let human drone experts play the follower role, i.e. controlling the simulated drone through our simulator interface, following the instructions and asking questions if they cannot find the destination area. 
When the experts stop the current navigation, they can either enter questions into a chatbox, claim the destination with a template sentence or reject the instruction for bad quality. If the destination is falsely claimed, the simulator will generate an auto-hint to let the follower ask some questions. For questions asked, AMT workers will provide further instructions accordingly based on given navigation information and dialog history. Then, the same drone experts will continue playing the follower role and again asking questions when necessary. We iterate the process until the destination is successfully reached and claimed by the follower.

\paragraph{Success Condition}
\label{success_condition}
The navigation trajectory is successful only when the destination is reached at the time the follower claims it. We determine that the destination is reached in view area $u_j$ by checking the center $c_{j}$ and computing the Intersection over Union (IoU) between $u_{j}$ and $Des$. If $c_j$ is inside $Des$ and the IoU of $u_{j}$ and $Des$ is larger than 0.4, the destination is regarded in $u_j$.




\subsection{Data Analysis}

Our AVDN dataset includes 3,064 aerial navigation trajectories, each with multi-round natural language dialog. There are two rounds of dialog on average per trajectory, where the number of dialog rounds in a trajectory equals to the maximum time step $M$. The most frequent words are shown in Figure \ref{words}. The recorded AVDN trajectory path length has an average of 287$m$, and its distribution is shown in Figure \ref{distr}. The trajectories and dialogs can be further separated into 6,269 sub-trajectories corresponding to the dialog rounds.

We split our dataset into \emph{training}, \emph{seen-validation}, \emph{unseen-validation}, and \emph{unseen-testing} sets, where \emph{seen} and \emph{unseen} sets are pre-separated by making sure the area locations of the visual scenes are over 100$km$ apart from each other. We show some statistical analysis across the dataset splits in Table~\ref{statistics}. The visual scenes in our dataset come from the xView dataset \cite{lam2018xview}, which covers both urban and rural scenes. The average covered area of the satellite images is 1.2$km^2$.


Rather than providing a target hint in the beginning as in \citet{thomason2020vision}, the destination must be inferred from the human instructions given by the commander.
For example, the commander may give a detailed description of the destination initially or write a rough instruction first and then describe the destination later in the dialog. We also find that there are two ways of describing the directions for navigation: egocentric direction description, such as ``turn right'', and allocentric direction description, such as ``turn south''. By filtering and categorizing words related to directions, we find that $82\%$ of the dialog rounds use egocentric direction description and $30\%$ of the dialog rounds include allocentric direction description. There are $17\%$ dialog rounds that have mixed direction deceptions, making the instruction complex. This opens a new challenge for developing a language understanding module that can ground both the egocentric and allocentric descriptions to navigation actions. 

\begin{table}[t]
\setlength\tabcolsep{0pt}
\setlength{\abovecaptionskip}{0.1cm}
\centering
\resizebox{\columnwidth}{!}{
        \begin{tabular}{lrrrrrrr}
        \toprule 
        \textbf{Split} 
        
        &   \textbf{\#dialogs} 
        & $\begin{array}{c}
            \textbf{\#words}
            \\
             \textbf{per dialog} 
                \end{array}$
        &  \textbf{\#areas}
        & $\begin{array}{c}
             \textbf{destination} \\
             \textbf{area-dim}
            \end{array}$
       & $\begin{array}{c}
            \textbf{\#sub-}
            \\
             \textbf{paths} 
            \end{array}$
       & $\begin{array}{c}
            \textbf{sub-path} 
            \\
             \textbf{length} 
            \end{array}$
        \\
        \midrule 
        Training&   2221 & 90 & 350 & 126m & 4591 & 145m \\
        Seen-val&    197 & 79 & 197 & 120m & 370 & 148m\\
        Unseen-val&     214 & 83 & 30 & 131m  & 411 &144m \\
        Unseen-test&    432 & 91 & 65 & 117m  & 897 & 142m \\
        \bottomrule
        \end{tabular}
    }
    \caption{Dataset statistics. \#dialogs is the number of dialogs, and \#words per dialog is the average number of words in each dialog. \#areas refers to the number of non-overlapped satellite images used. Destination area-dim is the average dimension of the sampled destination areas. \#sub-paths is the number of sub-trajectories where each sub-trajectory corresponds to one round of dialog. Sub-path length is the average sub-trajectory length.}
    \label{statistics}
\end{table}

\begin{figure}[t]
\centering
\setlength{\abovecaptionskip}{0.1cm}
\begin{minipage}[t]{0.2\textwidth}
\subfloat[\small{Frequent words}]{
\includegraphics[height = 2.3cm, width = 3.5cm]{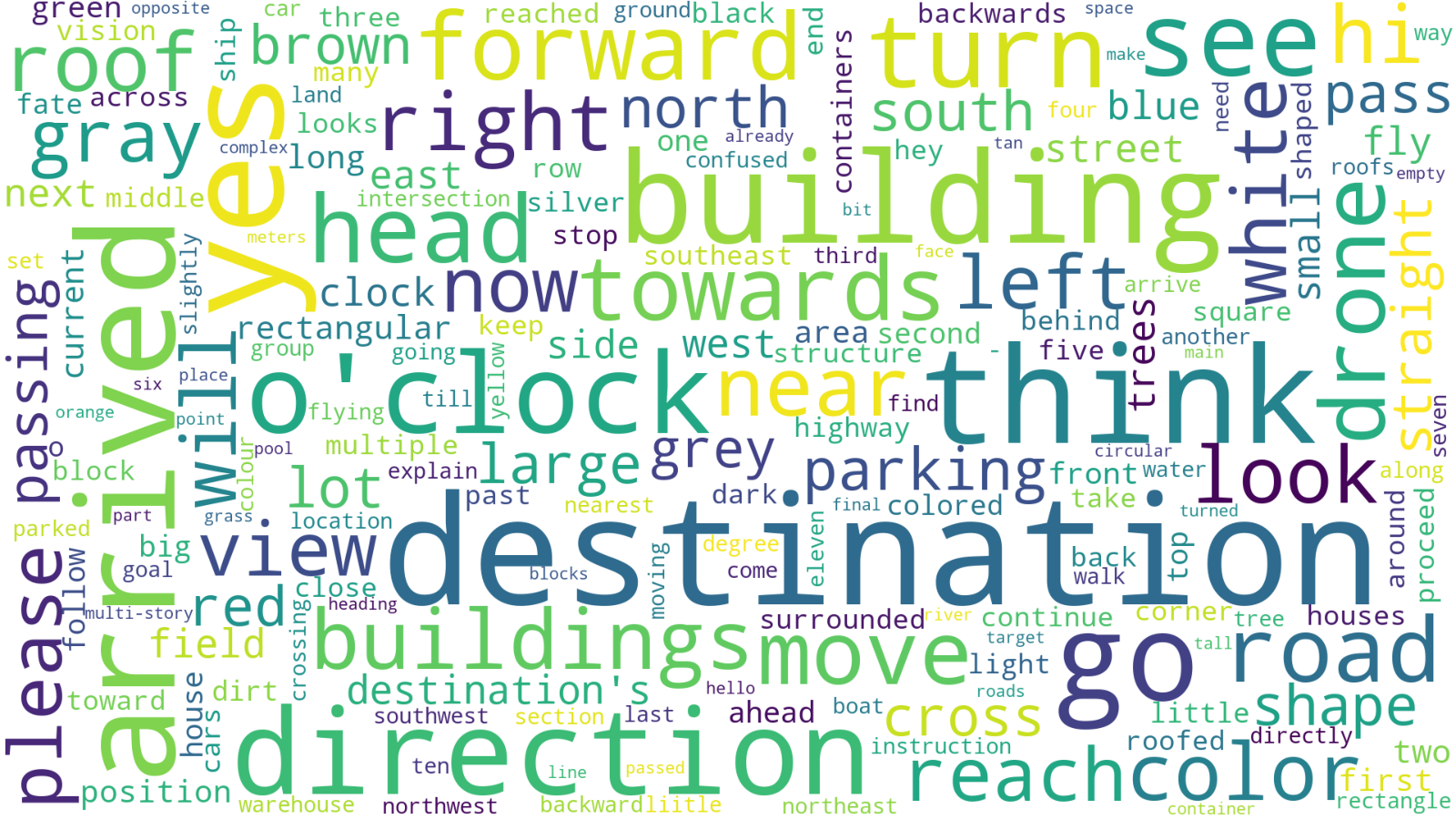}
\label{words}}
\end{minipage}
\hspace{0.2cm}
\begin{minipage}[t]{0.25\textwidth}
\subfloat[\small{Path length distribution}]{
\includegraphics[height = 2.3cm,  width = 4.4cm]{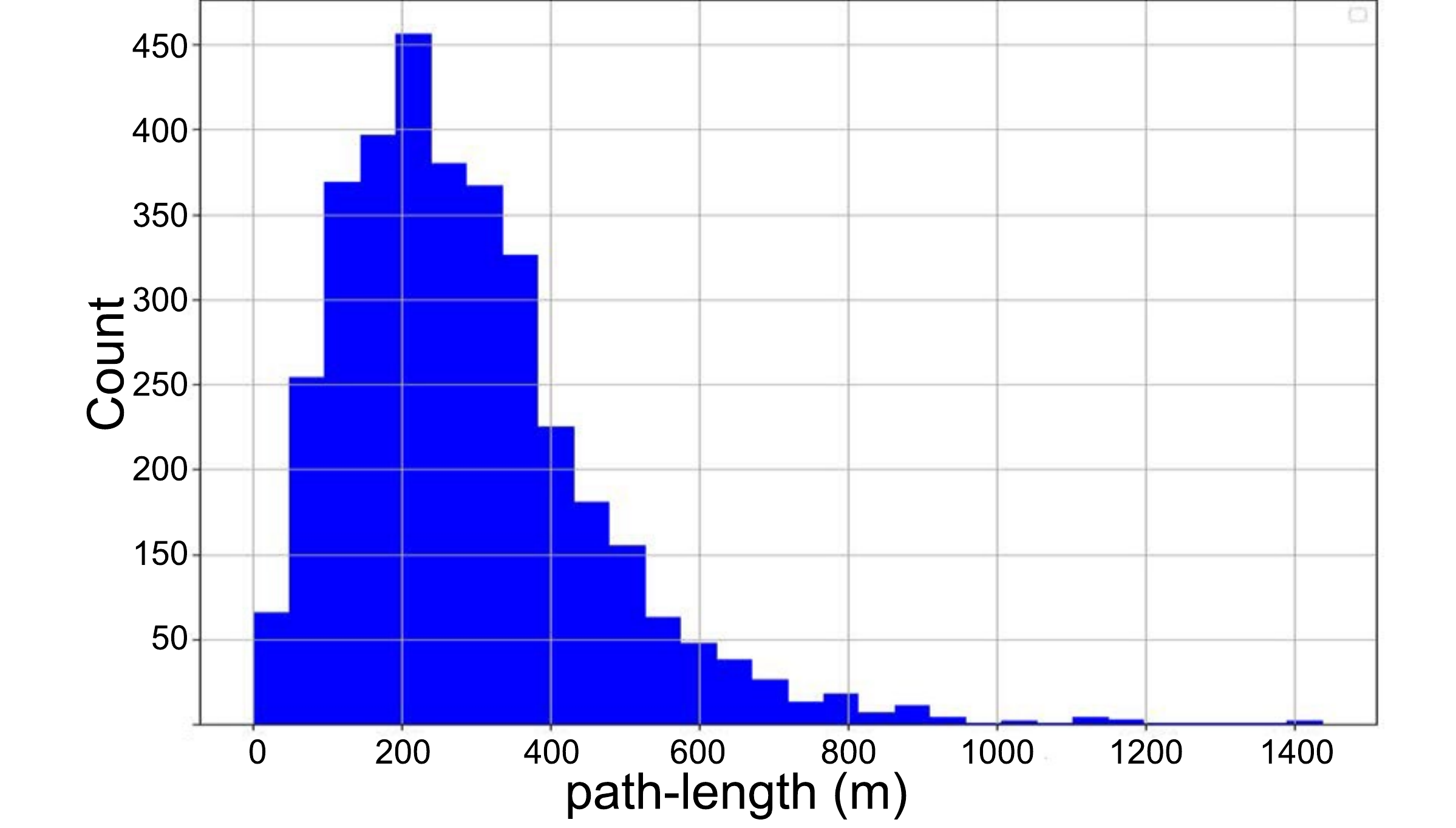}
\label{distr}}
\end{minipage}

\caption{(a) displays the frequent words that appear in the dialogs and (b) shows the path length distribution of our AVDN dataset.
}
\end{figure}

\section{Task}
Following indoor dialog navigation~\cite{thomason2020vision,kim2021ndh}, we introduce an Aerial Navigation from Dialog History (ANDH) task and an Aerial Navigation from Full Dialog History (ANDH-Full) task based on our AVDN dataset and simulator.

\subsection{Aerial Navigation from Dialog History} 

The goal of the task is to let the agent predict aerial navigation actions that lead to goal areas $G$, following the instructions in the dialog history. Specifically, to predict one action $\hat{a}_j$ of an action sequence between navigation time step $T_i$ and $T_{i+1}$, the inputs are dialogs from navigation time step $0$ to $T_i$ and images from a sequence of view areas $<\hat{u}_0, \hat{u}_1, \dots, \hat{u}_{j-1}>$. A new view area $\hat{u}_j$ will be generated after $\hat{a}_j$ takes place.\footnote{$\hat{u}_0$ is known as it is the initial view area at time step $T_i$.} The goal area $G$ depends on the current navigation time step,
\begin{equation}
\setlength{\abovedisplayskip}{1pt} \setlength{\abovedisplayshortskip}{1pt}
\setlength{\belowdisplayskip}{0pt} \setlength{\belowdisplayshortskip}{0pt}
    \begin{aligned}
    G =
    \begin{cases}
     u^{T_{i+1}}_0   ,& \text{if } T_{i+1}\neq M\\
    Des,              & \text{otherwise}
\end{cases}, 
\end{aligned}
\end{equation} 
The predicted view area sequence will be recorded for evaluation with regard to the ground truth view area sequence $<u^{T_{i}}_0, \dots, u^{T_{i}}_{N_{T_i}}>$.

\subsection{Aerial Navigation from Full Dialog History}

Compared with the ANDH task, the major difference of the ANDH-Full task is that it adopts the complete dialog history from navigation time step $T = 0, 1, \dots, M$ as input. With the full dialog and visual observation, the agent needs to predict the full navigation trajectory from the starting view area $u^0_{0}$ to the destination area $Des$. ANDH-Full provides complete supervision for agents on a navigation trajectory with a more precise destination description and includes longer utterances and more complex vision grounding challenges.

\subsection{Evaluation}
Since the agent in both tasks, ANDH and ANDH-Full, needs to generate predicted view area sequences, the evaluation metrics for both tasks are the same. In the evaluation, the center points of every view area are connected to form the navigation trajectory, and the last view area is used to determine whether the predicted navigation successfully leads to the destination area. The predicted navigation is successful if the IoU between the predicted final view area and the destination area is greater than 0.4. We apply several metrics for evaluation.

\noindent\textbf{Success Rate (SR)}: the number of the predicted trajectory being regarded as successful, i.e., the final view area of the predicted trajectory satisfies the IoU requirement, over the number of total trajectories predicted. 

\noindent\textbf{Success weighted by inverse Path Length (SPL)} \cite{anderson2018vision}: measuring the Success Rate weighted by the total length of the navigation trajectory. 

\noindent\textbf{Goal Progress (GP)} \cite{thomason2020vision}: evaluating the distance of the progress made towards the destination area. It is computed as the Euclidean distance of the trajectory, deducted by the remaining distance from the center of the predicted final view area $\hat{c}_{N}$ to the center of goal area $G$. 



\section{Model} \label{model}
\begin{figure*}[t]
\setlength{\abovecaptionskip}{-0.1cm}
    \centering
    \includegraphics[width = 0.9\linewidth]{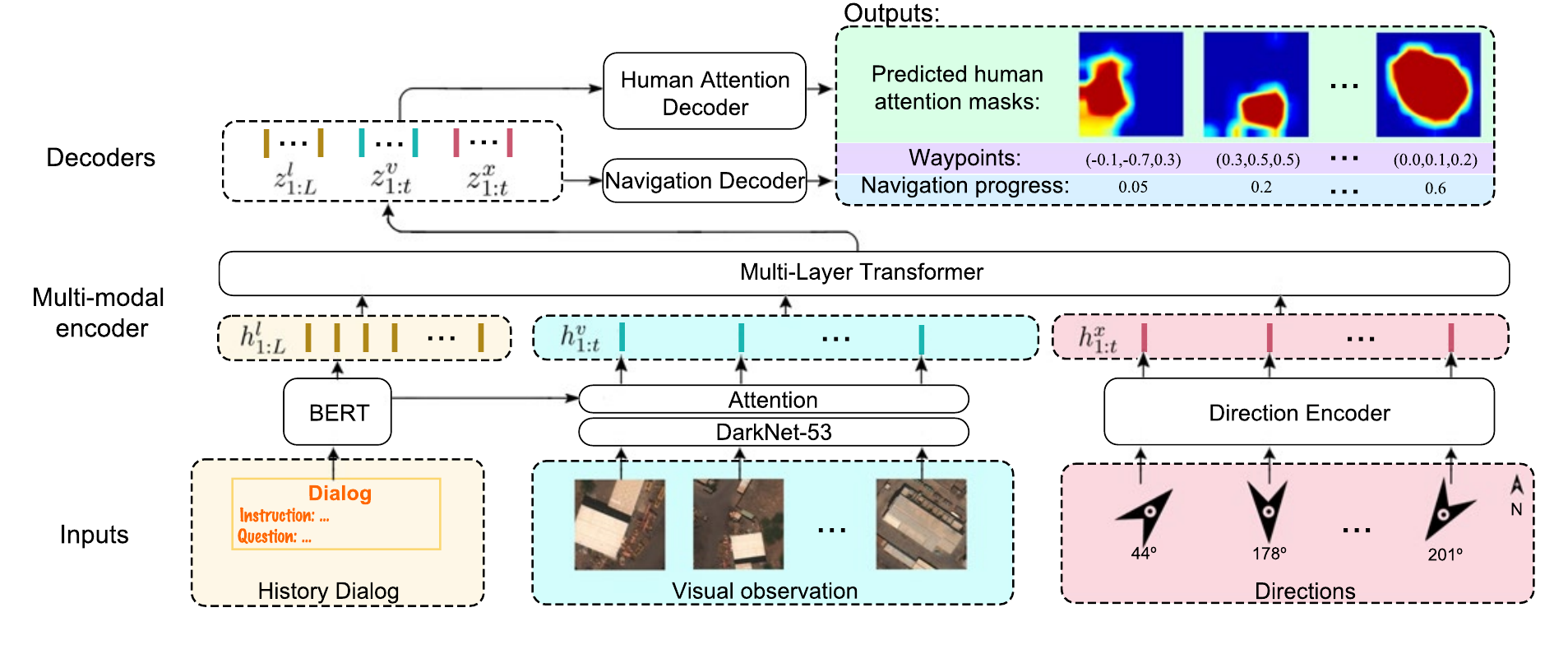}
    \caption{Our Human Attention Aided (HAA) model.
    The output of the model will interact with our simulator for generating the input for next time step.
    }
    \label{fig:structure}
\vspace{-2mm}
\end{figure*}

We proposed a Human Attention Aided (HAA-Transformer) model for the ANDH and ANDH-Full tasks as shown in Figure~\ref{fig:structure}, where it takes as input multimodal information and generates multimodal predictions, including human attention prediction and navigation prediction. 

\noindent\textbf{Multimodal Encoding~}
The input has three modalities, the drone's direction, images from the drone's visual observation, and history dialogs. At the start of a prediction series, our model uses a BERT encoder \cite{devlin2018bert} to get the language embeddings of the input dialog history, $ h^l_{1:L}$, where special language tokens such as [INS] and [QUE] are added in front of each instruction and question in the dialog. Then, at every time step, all previous drone directions and images from the drone's visual observation are input to the model. A fully connected direction encoder is used to generate direction embeddings $h^x_{1:t}$ and an xView-pretrained Darknet-53\footnote{\footnotesize{\url{https://github.com/ultralytics/xview-yolov3}}} \cite{redmon2018yolov3} with an attention module is used to extract and flatten the visual features to get visual embeddings $h^v_{1:t}$. Finally, similar to the Episodic Transformer \cite{pashevich2021episodic}, all embeddings from the languages, images and directions, are concatenated and input into a multimodal transformer ($F_{MT}$) to produce output multimodal embeddings $\{z^l_{1:L}, z^v_{1:t}, z^x_{1:t}\}$ as in Equation~\ref{eq_1}.
\begin{equation}
\label{eq_1}
\setlength{\abovedisplayskip}{1pt} \setlength{\abovedisplayshortskip}{1pt}
    \begin{aligned}
    \{z^l_{1:L}, z^v_{1:t}, z^x_{1:t}\} 
    =  F_{MT}(\{h^l_{1:L}, h^v_{1:t}, h^x_{1:t}\})
\end{aligned}
\end{equation}

\noindent\textbf{Navigation Prediction and Waypoint Control }
The navigation outputs from our model come from a fully connected navigation decoder ($F_{ND}$) taking as input the transformer's output embeddings $\{z^l_{1:L}, z^v_{1:t}, z^x_{1:t}\}$ and generating predicted waypoint actions $\hat{w}$ and predicted navigation progress $\hat{g}$ as in Equation~\ref{eq2}.
\begin{equation}
\label{eq2}
\setlength{\belowdisplayskip}{2pt} \setlength{\belowdisplayshortskip}{2pt}
\setlength{\abovedisplayskip}{2pt} \setlength{\abovedisplayshortskip}{2pt}
    \begin{aligned}
    (\hat{w}, \hat{g}) = F_{ND}( \{z^l_{1:L}, z^v_{1:t}, z^x_{1:t}\})
\end{aligned}
\end{equation} 
The predicted waypoint action $\hat{w}$ is a 3-D coordinate $(\hat{x}, \hat{y}, \hat{h})$, where $(\hat{x},\hat{y})$ corresponds to an  position in the current view area $u$ and $\hat{h}$ corresponds to an altitude. The predicted waypoint also controls the drone's direction, where the direction is kept towards the direction of movement. Therefore, $\hat{w}$ controls the drone's movement, and as a result, the center, width and rotation of the next view area center are determined by $\hat{w}$. As for the navigation progress prediction $\hat{g}$, it is to generate a one-dimension navigation progress indicator for deciding when to stop \cite{xiang2019not}. If the predicted navigation progress is larger than a threshold, the drone navigation will be ended without executing the predicted waypoint action. 

\noindent\textbf{Human Attention Prediction}
A human attention decoder is proposed to predict the human attention mask using the output embeddings, $z^v_{1:t}$, from the multi-layer transformer that corresponds to the visual inputs. We build the decoder based on \citet{he2019understanding}, where the input to the decoder will be decoded to an $8*8$ representation through a fully connected layer and then linearly interpolated to a mask with the same shape as the input image. The greater the values in the mask means more likely the human follower attends the corresponding pixels. 

\noindent\textbf{Training}
We first train our HAA-Transformer model on the ANDH task and then fine-tuned it on the ANDH-Full task because the ANDH task is relatively easier with a shorter path length. For each task, we conduct the training alternately in teacher-forcing \cite{williams1989learning} and student-forcing modes, where the main difference is whether the model interacts with the simulator using ground truth actions or the predicted actions. Our model is trained with a sum of losses from both navigation prediction and human attention prediction. 
First, the predicted waypoint action $\hat{w}$ and predicted navigation progress $\hat{g}$ are trained with Mean Square Error (MSE) loss, supervised by the ground truth $w$ and $g$ computed based on the recorded trajectory in our dataset. The navigation prediction loss ($L_{nav}$) is shown in Equation~\ref{eq3}, where $\it{Rot}(.)$ is computing the rotation change as a result of the waypoint action.
\begin{equation}
\label{eq3}
\setlength{\abovedisplayskip}{2pt} \setlength{\abovedisplayshortskip}{2pt}
\setlength{\belowdisplayskip}{2pt} \setlength{\belowdisplayshortskip}{2pt}
    \begin{aligned}
    L_{nav}
    = &\it{MSE}(\it{Rot}(\hat{w}),\it{Rot} (w)) \\
    +& \it{MSE}(\hat{w}, w) +  \it{MSE}(\hat{g}, g)
\end{aligned}
\end{equation}
Second, for human attention prediction training, we apply the modified Normalized Scanpath Score loss (NSS) \cite{he2019understanding}. Given a predicted human attention mask $P$ and a ground-truth human attention mask $Q$,
\begin{equation}
\setlength{\abovedisplayskip}{2pt} \setlength{\abovedisplayshortskip}{2pt}
\setlength{\belowdisplayskip}{2pt} \setlength{\belowdisplayshortskip}{2pt}
    \begin{aligned}
    &N S S\left(P, Q\right)=\frac{1}{N} \sum_{i} \overline{P_{i}} \times Q_{i}, \\
&\text { where } N=\sum_{i} Q_{i} \text { and } \bar{P}=\frac{P-\mu(P)}{\sigma(P)}
\end{aligned}
\end{equation}
Since human attention may not exist in certain view areas, the human attention loss is only computed for view areas with recorded human attention. 

\begin{table*}[t]
\setlength{\abovecaptionskip}{0.1cm}
    \centering
    \setlength\tabcolsep{5pt}
\resizebox{1\textwidth}{!}{
\begin{tabular}{l|rrr|rrr|rrr|rrr|rrr|rrr} 
\toprule

& \multicolumn{9}{c|}{\textbf { ANDH }} & \multicolumn{9}{c}{\textbf { ANDH-Full }}  \\
\cmidrule(lr){2-10} \cmidrule(lr){11-19}
& \multicolumn{3}{c}{\textbf { Seen Validation }} & \multicolumn{3}{|c}{\textbf { Unseen Validation }}& \multicolumn{3}{|c}{\textbf { Unseen Testing }} & \multicolumn{3}{|c}{\textbf { Seen Validation }} & \multicolumn{3}{|c}{\textbf { Unseen Validation }}& \multicolumn{3}{|c}{\textbf { Unseen Testing }} \\


\textbf{Model} & \multicolumn{1}{r}{
\text {SPL}$\uparrow$
} & \multicolumn{1}{r}{
\text {SR}$\uparrow$
} & \multicolumn{1}{r}{
\text {GP}$\uparrow$
}& \multicolumn{1}{|r}{
\text {SPL}$\uparrow$
} & \multicolumn{1}{r}{
\text {SR}$\uparrow$
} & \multicolumn{1}{r}{
\text {GP}$\uparrow$
} & \multicolumn{1}{|r}{
\text {SPL}$\uparrow$
} & \multicolumn{1}{r}{
\text {SR}$\uparrow$
} & \multicolumn{1}{r}{
\text {GP}$\uparrow$
} & \multicolumn{1}{|r}{
\text {SPL}$\uparrow$
} & \multicolumn{1}{r}{
\text {SR}$\uparrow$ 
} & \multicolumn{1}{r}{
\text {GP}$\uparrow$
}
 & \multicolumn{1}{|r}{
\text {SPL}$\uparrow$
} & \multicolumn{1}{r}{
\text {SR}$\uparrow$
} & \multicolumn{1}{r}{
\text {GP}$\uparrow$
} & \multicolumn{1}{|r}{
\text {SPL}$\uparrow$
} & \multicolumn{1}{r}{
\text {SR}$\uparrow$
} & \multicolumn{1}{r}{
\text {GP}$\uparrow$
}
\\
\midrule
$\begin{array}{l}
\text {Random}\\
\end{array}$  & 0.5 & 1.6 & -84.1  & 0.2 & 1.0 & -81.4 & 0.5  & 1.1  & -86.6  & 0.0 & 0.0 & -176.6 & 0.0 & 0.0 & -162.4 & 0.2 & 0.1 & -158.4\\
$\begin{array}{l}
\text{Vision-only E.T.}\\
\end{array}$ & 3.1 & 3.8 & 0.2  & 6.5 & 7.3 & 0.2 & 3.2 & 3.9 & 0.2 & 0.0 & 0.0 & -1.1  & 1.9 & 1.9 & -5.2 & 0.2 & 0.2 & -1.6 \\
$\begin{array}{l}
\text {Language-only E.T.}\\
\end{array}$ & 9.2 & 11.6 & 48.9  & 13.9 & 17.0 & 50.6 & 9.7 & 12.7 & 49.1 &    1.4 & 2.0 & 54.8  & 2.3 & 3.7 & 54.5 & 1.8 & 2.2 & 58.2\\
\midrule
$\begin{array}{l}
\text {E.T.}\\
\end{array}$ & 12.1 & 14.1 & 50.1 & 14.3 & 16.6 & 51.9 & 11.3 & 13.3  & 51.7  &      2.2 & 3.1 & 51.3 & 2.5 & 3.7 & 48.9 & 1.9 & 2.8 & 60.7  \\
$\begin{array}{l}
\text {HAA-Transformer}\\
\end{array}$ & \textbf{14.7} & \textbf{17.3} & \textbf{56.3} & \textbf{16.5} & \textbf{20.4} &\textbf{55.2} & \textbf{12.9} & \textbf{15.7}  & \textbf{53.7}      & \textbf{3.7} & \textbf{5.1} & \textbf{54.6}  & \textbf{3.2} & \textbf{4.7} & \textbf{50.9} & \textbf{4.1} & \textbf{6.3}   & \textbf{63.2}  \\
\midrule
$\begin{array}{l}
\text {LSTM}\\
\end{array}$ & 9.0 & 10.3 & 31.9 & 13.3 & 14.1 & 35.9 & 9.7 & 10.8  & 40.4  &     1.0 & 1.0 & 43.8 & 3.2 & 3.7 & 48.7 & 1.8 & 1.9 & 56.4  \\
$\begin{array}{l}
\text {HAA-LSTM}\\
\end{array}$ & \textbf{11.6} & \textbf{13.0} & \textbf{50.3} & \textbf{18.3} & \textbf{20.0} & \textbf{54.4} & \textbf{12.6} & \textbf{14.1}  & \textbf{54.6}  &     \textbf{3.8} & \textbf{4.1} &  \textbf{52.2} & \textbf{3.4} & 3.7 & \textbf{56.1} & \textbf{1.9} & \textbf{2.6}   & \textbf{66.5}  \\
\bottomrule

\end{tabular}
}
\caption{Main results on both ANDH and ANDH-Full tasks including ablation results on human attention prediction training. Both the Human Attention Aided Multi-modal LSTM (HAA-LSTM) model and our HAA-Transformer model are benefited from the human attention prediction training based on the performance comparison.\label{tab:experiment_0}}
\vspace{-5mm}
\end{table*}

\section{Results}
We conduct experiments to study our AVDN dataset and our HAA-Transformer model on the ANDH and ANDH-Full tasks.

\noindent\textbf{Results on the ANDH task and ANDH-Full task}\label{impact_exp} 
As shown in Table~\ref{tab:experiment_0}, we evaluate our HAA-Transformer model along with multiple baseline models on both ANDH and ANDH-Full tasks. We first create a multimodal Episodic Transformer (E.T.) model \cite{pashevich2021episodic} by removing the human attention decoder from our HAA-Transformer, and then build vision-only and language-only uni-modal models by ablating on the multimodal E.T. model. For uni-modal models, direction inputs are maintained while either vision input or language input is discarded. A multimodal LSTM-based model is also included as a sequence-to-sequence baseline model, which has the same input and output as the multimodal E.T.model. All models, including our HAA-Transformer model are trained with random initialization. The batch size is 4 for the ANDH task, while for the ANDH-Full task. Based on the result, our HAA-Transformer model outperforms the baseline models in both tasks by a large margin. Also, compared with uni-modal baseline models and a random model outputting random waypoint actions, the multimodal E.T. model achieves overall higher performance, which indicates the importance of learning multimodal information in order to succeed in the ANDH task. Last but not least, we find that the language-only uni-modal model achieves much better performance than the vision-only uni-modal model showing that the language instructions play a more important role in guiding the navigation in our AVDN dataset.


\noindent\textbf{Impact of Human Attention Prediction Training}
We then evaluate the impact of human attention prediction training for multimodal learning by ablation not only on our HAA-Transformer model but also on a Human Attention Aided Multimodal LSTM-based (HAA-LSTM) model developed by adding the human attention decoder module to the multimodal LSTM-based model (detailed in Appendix \ref{lstm_model}). We apply the same human attention prediction training process and training loss as in our HAA-Transformer model. As the result shown in Table.~\ref{tab:experiment_0}, we find that the human attention prediction training significantly boosts both transformer-based and LSTM-based models across all evaluated metrics.

\begin{figure}
    \centering
    \setlength{\abovedisplayskip}{0pt}
    \setlength{\abovecaptionskip}{-0.1cm}
    \includegraphics[width=1\linewidth]{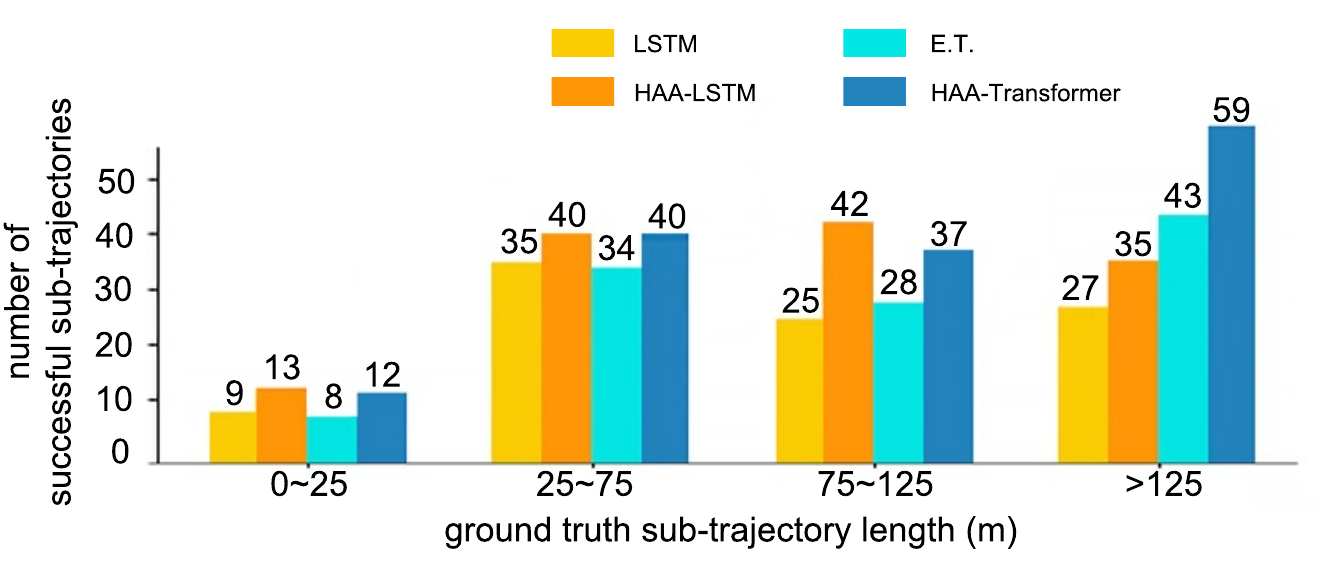}
    \caption{The impact of human attention prediction training on the success of trajectories of different lengths. Human attention prediction significantly improves navigation performance for longer trajectories.}
    \label{fig:exp_ab}
\end{figure}

We further evaluate the benefit of human attention prediction training on different trajectory lengths. The sub-trajectories in the validation set for ANDH task are split into four subsets based on the ground truth length. In Figure \ref{fig:exp_ab}, we compare the number of successful sub-trajectory in different subsets among models with and without human attention prediction training. As a result, both our HAA-Transformer model and the HAA-LSTM model achieves significant performance improvements for subsets of longer trajectory. It leads to the conclusion that human attention prediction training benefits navigation prediction, especially for long trajectories for both two models that are based on LSTM and Transformer. 

Besides improving task performance, human attention prediction also benefits the interpretability of the model by generating visualizable attention predictions paired with navigation predictions. We evaluate the human attention prediction result using the Normalized Scanpath Saliency (NSS) score, which measures the normalized saliency prediction at the ground truth human attention. Our HAA-Transformer model receives NSS scores of 0.84, 0.62 and 0.68, respectively, in seen validation, unseen validation, and test set, indicating the human attention prediction is effective.

\noindent\textbf{Comparison for Different Input Dialog Length} Comparing with the ANDH task, the ANDH-Full task requires the model to predict actions that correspond to longer dialogs with more dialog rounds. As a result, more challenges are involved and longer training time is needed compared with the results in the ANDH task. During training, we add a prompt of the drone's direction that corresponds to the dialog, e.g., ``when facing east'' to clarify the instructions in dialogs that happened in different time steps, especially when egocentric direction descriptions exist. In Table~\ref{tab:experiment_2}, we show our HAA-Transformer model's performance on trajectories with different dialog lengths, i.e. different numbers of dialog rounds, and we find the model's SR and SPL are diminished for trajectories with the number of dialog rounds less or greater than average, where they either containing too less or too much information. It shows a big room for improvement in understanding dialog with various lengths.

\begin{table}[t]
    \centering
\setlength{\abovecaptionskip}{0.1cm}
\setlength\tabcolsep{4pt}
\resizebox{0.7\columnwidth}{!}{
\begin{tabular}{l|cc|cc|cc} 
\toprule

& \multicolumn{2}{c|}{$\begin{array}{c}
     \textbf{Seen}  \\
     \textbf{Validation} 
\end{array}$} & \multicolumn{2}{c|}{$\begin{array}{c}
     \textbf{Unseen}  \\
     \textbf{Validation} 
\end{array}$} & \multicolumn{2}{c}{$\begin{array}{c}
     \textbf{Unseen}  \\
     \textbf{Testing} 
\end{array}$}  \\

 \cmidrule(lr){2-3} 
 \cmidrule(lr){4-5} 
 \cmidrule(lr){6-7} 
 
\multicolumn{1}{c}{\textbf{Dialog rounds}} 
&\multicolumn{1}{|r}{
\text {SPL}$\uparrow$} 
& \multicolumn{1}{r}{
\text {SR}$\uparrow$
} & \multicolumn{1}{|r}{
\text {SPL}$\uparrow$
} & \multicolumn{1}{r}{
\text {SR}$\uparrow$
} & \multicolumn{1}{|r}{
\text {SPL}$\uparrow$
} & \multicolumn{1}{r}{
\text {SR}$\uparrow$
} 
\\
\midrule
$\begin{array}{l}
    \text{1}\\
\end{array}$   & 3.7 & 4.7  & 1.7 & 2.2 & 4.3 & 7.0   \\
$\begin{array}{l}
    \text {2}\\
\end{array}$   & \textbf{4.3} & \textbf{6.2}  & \textbf{4.8} & \textbf{7.3} & \textbf{6.8} & \textbf{10.3}    \\
$\begin{array}{l}
    \geq\text {3}\\
\end{array}$   & 2.7 & 4.3  & 3.8 & 6.1 & 1.6 & 2.0    \\
\midrule
$\begin{array}{l}
    \text {All}\\
\end{array}$   & 3.7 & 5.1 & 3.2 & 4.7 & 4.1 & 6.3  \\

\bottomrule
\end{tabular}
}
\caption{Result of our HAA-Transformer on ANDH-Full task regarding different dialog lengths. The more rounds in dialog, the longer the trajectory is and the more challenging the task is. \label{tab:experiment_2} }
\vspace{-3mm}
\end{table}

\section{Conclusion} 
In this work, we introduce a dataset and a simulator for Aerial Vision-and-Language Navigation (AVDN). Challenging tasks are proposed based on our dataset focusing on navigation. A Human Attention Aided Multimodal Transformer (HAA-Transformer) model is designed for both tasks. Our work provides the possibilities for further studies to develop stronger models on AVDN that not only focus on navigation prediction but also on question generation. Furthermore, based on our results, future works may investigate using human attention prediction training to help solve VLN problems.


\section*{Limitation}
This work proposed a dataset, a simulator, tasks, and models for Aerial Vision-and-Language Navigation. Since satellite images are needed to simulate the drone's observation, risks of privacy leaking may exist. By using the open-source satellite dataset xView \cite{lam2018xview}, we mitigate the risks while also being able to develop a simulator for training our model. Additionally, using satellite images for simulating
top-down visual observation of the drone introduces the shortcoming of having only 2D static
scenes while adopting the strength of the satellite images where rich labels and visual features are included.

\section*{Broader Impact}
We recognize the potential ethical problems during the dataset collection, where human annotators are involved. The data collection of this project is classified as exempt by Human Subject Committee vis IRB protocols. As a result, we utilized the Amazon Mechanical Turk (AMT) website to find workers willing to participate in the project. With AMT, our data collection is constrained by legal terms, and the data collection protocol is under AMT's approval. The agreement signed by both requesters and workers on AMT also ensures a transparent and fair data annotation process and that privacy is well protected.

\bibliography{acl_latex}
\bibliographystyle{acl_natbib}
\appendix

\newpage

\section{HAA-Transformer Model Details}
There are around 120m parameters in our HAA-Transformer model. Our model uses a $\text{BERT}_{\text{BASE}}$ encoder \cite{devlin2018bert} with pretrained weights that open-sourced on Hugging Face \cite{wolf-etal-2020-transformers} to extract language feature of the input dialog history. For ANDH task, We extract two sets of language embeddings in ANDH task, where the input is either all the previous and current dialog rounds, or only the current dialog round for the target sub-trajectory. The language embeddings that include all previous dialog are used to attend to the image feature extracted by DarkNet-53 and flatten the feature to only 768 long per frame. The other with only current dialog is passed to the multi-model encoder. Whereas in ANDH-Full task, since the agent starts at an initial position with no previous dialog, only one set of language embeddings is extracted and used.

The attention modules that are used in our HAA-Transformer model and the HAA-LSTM model have the same structure. They generate soft attention based on dot-product attention mechanism. The inputs are context features and attention features. There is a fully connected layer before the output of the attention module. The context features attended by the attention features are concatenated with the attention features to become the input of the fully connected layer, and the output will be the attention module's output which has the same shape as the attention features.

\subsection{Navigation Progress Prediction}
\label{stop}
As for the navigation progress prediction, we adopt the idea of L2Stop \cite{xiang2019not} and create a navigation progress predictor to help decide when to stop, which overcomes the problem that the model would fail to stop at the desired position. The navigation progress is trained with the supervision of IoU score of the current view area $\hat{u}_{i,j,k}$ and the destination area. When the IoU is larger than 0, it indicates the designation area is seen in $\hat{u}_{i,j,k}$ and the larger the IoU the closer the $\hat{u}_{i,j,k}$ to the $des_{i,j}$ . During the inference time, the predicted navigation stops when the generated navigation progress indicator is less than 0.5. 

\section{HAA-LSTM Model} \label{lstm_model}

\begin{figure}[t]
    \centering
    \includegraphics[width = \linewidth]{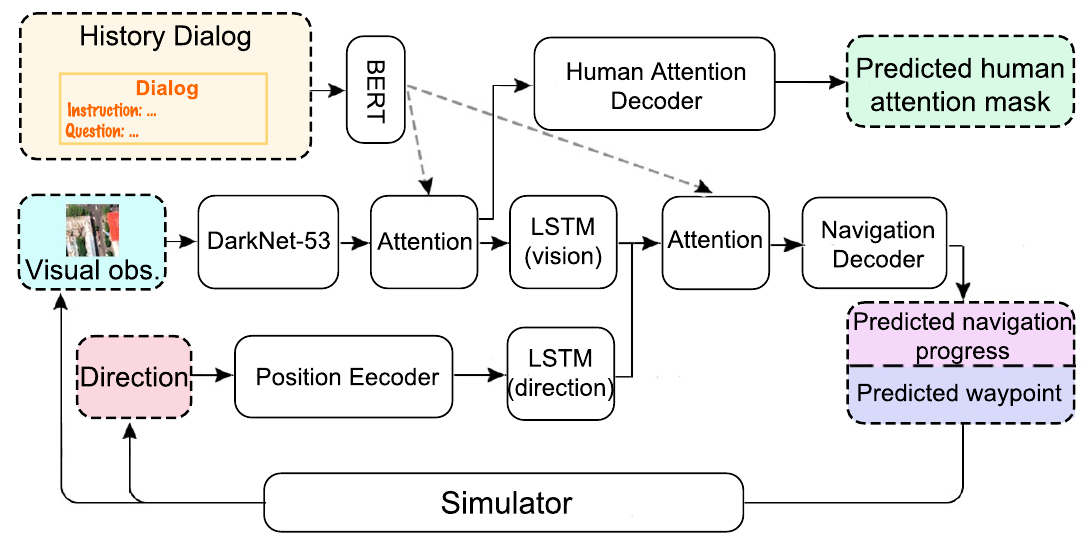}
    \caption{Human Attention Aided Multi-modal LSTM (HAA-LSTM) model which uses the same input and output as our HAA-Transformer.
    }
    \label{fig:structure_lstm}
\end{figure}

We also design a Human Attention Aided Multi-modal LSTM model for experiments in Section \ref{impact_exp} as shown in Figure \ref{fig:structure_lstm}, where it takes the same input and output as our HAA-Transformer model. We also add the same human attention decoder as in our HAA-Transformer model for human attention prediction training. The language embeddings, visual observation and direction embeddings are also extracted in the same way.

\section{Training Details}

We train all models on one Nvidia RTX A6000 graphic card.
We train all baseline models as well as both HAA-Transformer model and HAA-LSTM model for approximately 150k iterations on ANDH task with batch size being 4 and learning rate being 1e-5. For the ANDH-Full task, since it uses full dialog history as input, where more GPU RAM is needed, we use a batch size of 2 and learning rate of 5e-6 and train the model for 200k interactions which take about 48 hours.

\begin{figure}[t]
    \centering
        \begin{minipage}[b]{0.155\textwidth}
        \centering
        \subfloat[\tiny{Simulated drone.}]{
        \label{fig:justification_2}
        \includegraphics[width=0.9\linewidth, height=4.75cm]{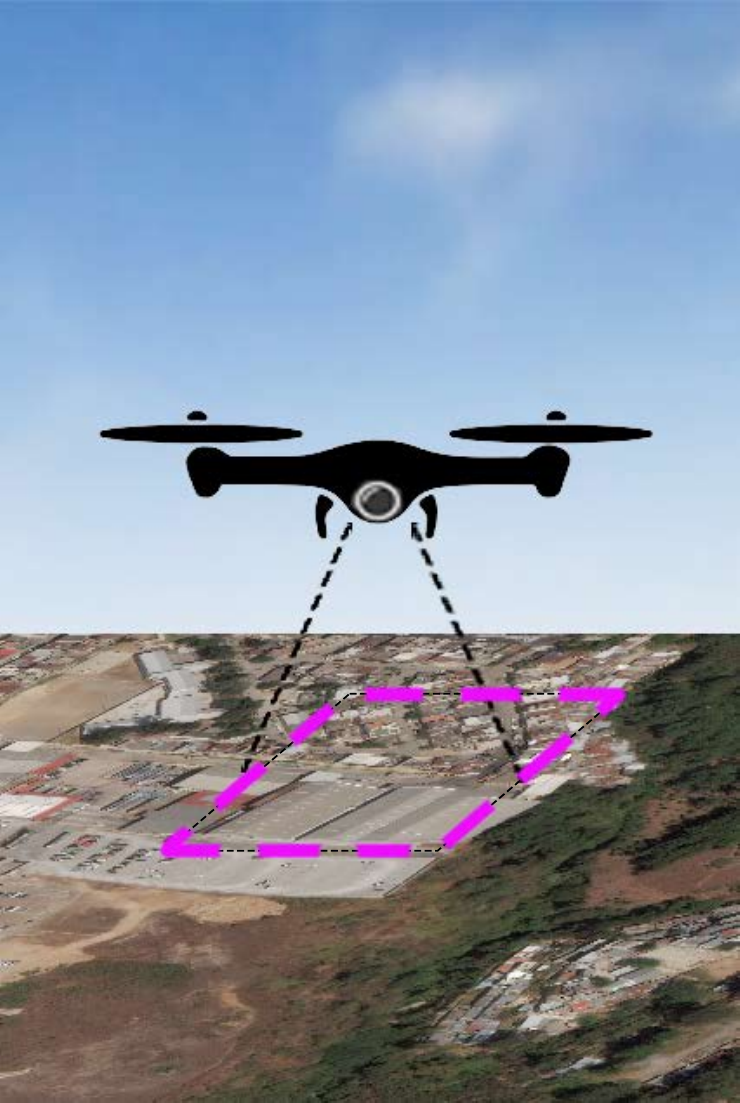}
        }
        \end{minipage}
        \hspace{-0.3cm}
        \begin{minipage}[b]{0.32\textwidth}
        \centering
        \subfloat[\tiny{Cropped satellite images.}]{
        \label{fig:justification_1}
        \includegraphics[width=2.0cm,height=2.0cm]{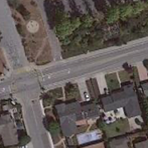}
        \includegraphics[width=2.0cm,height=2.0cm]{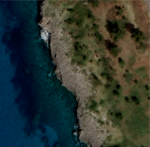}}
        \vspace{-0.1cm}
        
        \subfloat[\tiny{Real drone visual observation.}]{
        \label{fig:justification_0}
        \includegraphics[width=2.0cm,height=2.0cm]{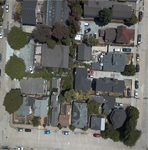}    
        \includegraphics[width=2.0cm,height=2.0cm]{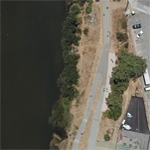}}
        \end{minipage}
        \label{fig:justification}
        \caption{(a) shows how simulated drone's visual observation is generated from satellite images in our simulator. We compare the simulated drone's visual observation from satellite images, (b), with images from a drone's onboard camera at about 200m above ground, (c).
        }

\end{figure}

\section{Simulator Details}

We design a simulator to simulate a drone flying with its onboard camera facing straight downward, as in Figure \ref{fig:justification_2}. The simulator uses satellite images from xView dataset \cite{lam2018xview} for the drone's visual observation, where the observation is square image patches cropped from the satellite image based on the drone's view area, as in Figure \ref{fig:justification_1}. We argue that by using satellite images, our simulator is capable of providing equally rich visual features as in the real world and some examples are shown in Figure \ref{fig:justification_0}. Additionally, since satellite images have boundaries that are not adjacent with each other, we prevent the drone's view area from moving out of boundary by automatically invalidate the drone's action that will lead to out-of-boundary view areas. Further more, for simplicity, we assume perfect control of the drone's movement, and therefore, the drone's current view area is determined by the previous drone's position and navigation action.

During the dataset collection, the follower controls the simulated drone through the simulator interface with keyboards. We defined 8 keys for the control with a total of four degrees of freedoms (DoFs), where there are 2 DoFs for horizontal movement, 1 DoF for altitude control, and 1 DoF for rotation control. Despite that our simulator environment is continuous, the control through the interface is discrete for an easier control experience. Every time a key is pressed, the simulated drone will move along the DoF for a fixed distance and the higher the simulated drone flies, the faster it moves with one press of the keyboard. Before the follower presses ESC key to stop the control, he/she can also generate the human attention data by using the mouse to left-click on the attended image region shown on the interface. After every left-click, a circle with a radius being 1/10 of the current view area width will become the attended region and be displayed on the interface. Also, a right-click on the circle will remove this region from the attention record.

\section{Dataset Details and Examples}

We provide some details about our dataset with related examples. Each example includes a dialog, sample drone's visual observation with human attention and navigation overviews.

\begin{figure}
    \centering
    \includegraphics[width = 0.45\textwidth]{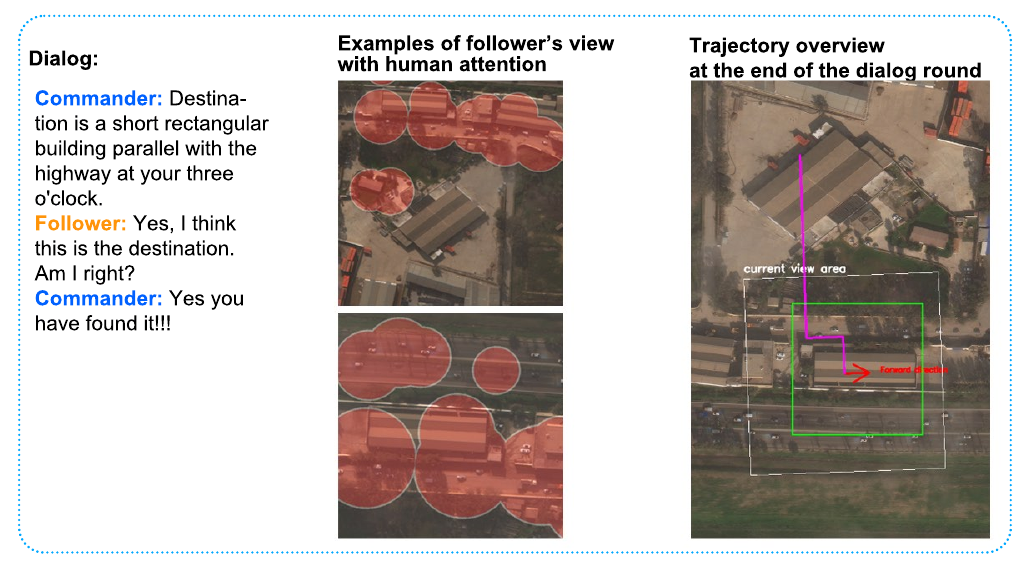}
    \caption{Example of a trajectory with one dialog round.}
    \label{fig:app_example_0}
\end{figure}

\begin{figure}
    \centering
    \includegraphics[width = 0.45\textwidth]{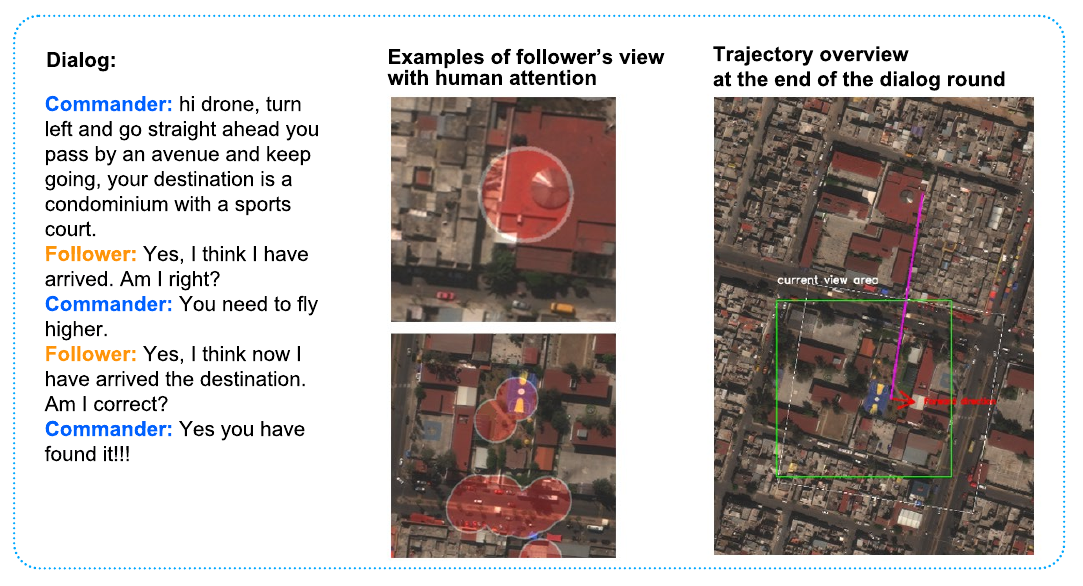}
    \caption{Example of a trajectory that includes auto-instruction about altitude adjustment. There is only one dialog round.}
    \label{fig:app_example_1}
\end{figure}

\begin{figure*}
    \centering
    \includegraphics[height = 5cm]{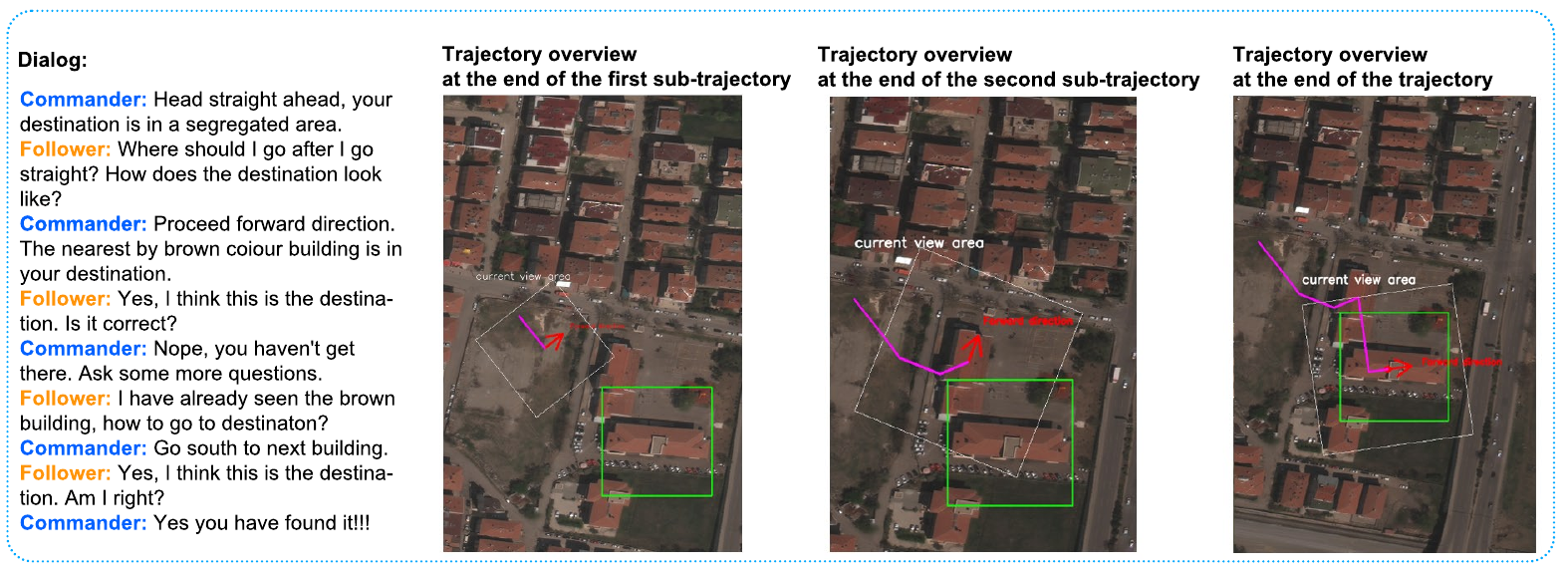}
    \caption{Example of a trajectory with three dialog rounds. There is an incorrect instruction in the second dialog round, where the destination should be described as the second nearest brown building rather than the nearest one. For this case, since the instruction is clear and can be followed by the follower, we treat it as an inevitable and acceptable type of instruction with mistakes and keep it in our dataset.}
    \label{fig:app_example_3}
\end{figure*}

\subsection{Human Attention}

We record the attention from the follower through our simulator interface when the follower is controlling the simulated drone. In each navigation trajectory collected, the attention are stored in a list where the order of the list is ignored, meaning that the attended areas either recorded earlier or later during the navigation will be retrieved together when using the human attention data. In this way, the human attention data becomes more accurate since the area that followers missed to attend in the current view area is likely to be included in the future time steps. Also, because the previously attended area is kept in later view areas, less effort is needed to annotate the attended areas. We find that 1/7 of the area on average is attended to in the recorded view areas $u_{i,j}$. 

\subsection{Dialog Structure}
The dialogs contained in our AVDN datset have a various number of rounds. Since the dialog rounds are split based on the data collection rounds, each dialog round contains only one instruction written by the commander. Figure \ref{fig:app_example_0} shows an example of a simple dialog with only one dialog round. However, when the follower can not follow the initial instruction to find the destination area, questions will be brought up, and therefore more dialog rounds will be introduced. Every dialog rounds start with the instruction from human commanders and could include one or more utterance from the follower, depending on if auto-instructions exist. We provide details about auto-instructions in the next sub-section. Also, when followers are writing the questions, we enable them to define some shortcut keys for frequently used general questions such as ``could you further explain it?'', ``where should I go?'', etc. To avoid templated dialogs, followers are forbidden to only use the shortcut for the question but need to incorporate their own language.

\subsection{Auto-instructions}
When the follower claims that the destination is reached, our simulator will check the navigation result automatically using the success condition described in Section \ref{success_condition}. Then, auto-instructions will be generated based on whether the destination area is reached successfully. Specifically, when the success condition is met, an auto-instruction of ``Yes, you have found it!!!'' will be added to the dialog as the end; if the destination is in the center of the view area, but the view area is either too large or too small, failing the success condition, the simulator will also provide auto-instructions asking the follower to adjust the drone's altitude and verify again if the success conditions are met or not, as shown is Figure \ref{fig:app_example_1}. 

\begin{figure}
    \centering
    \subfloat[Words from commander utterances]{\includegraphics[width=1\linewidth]{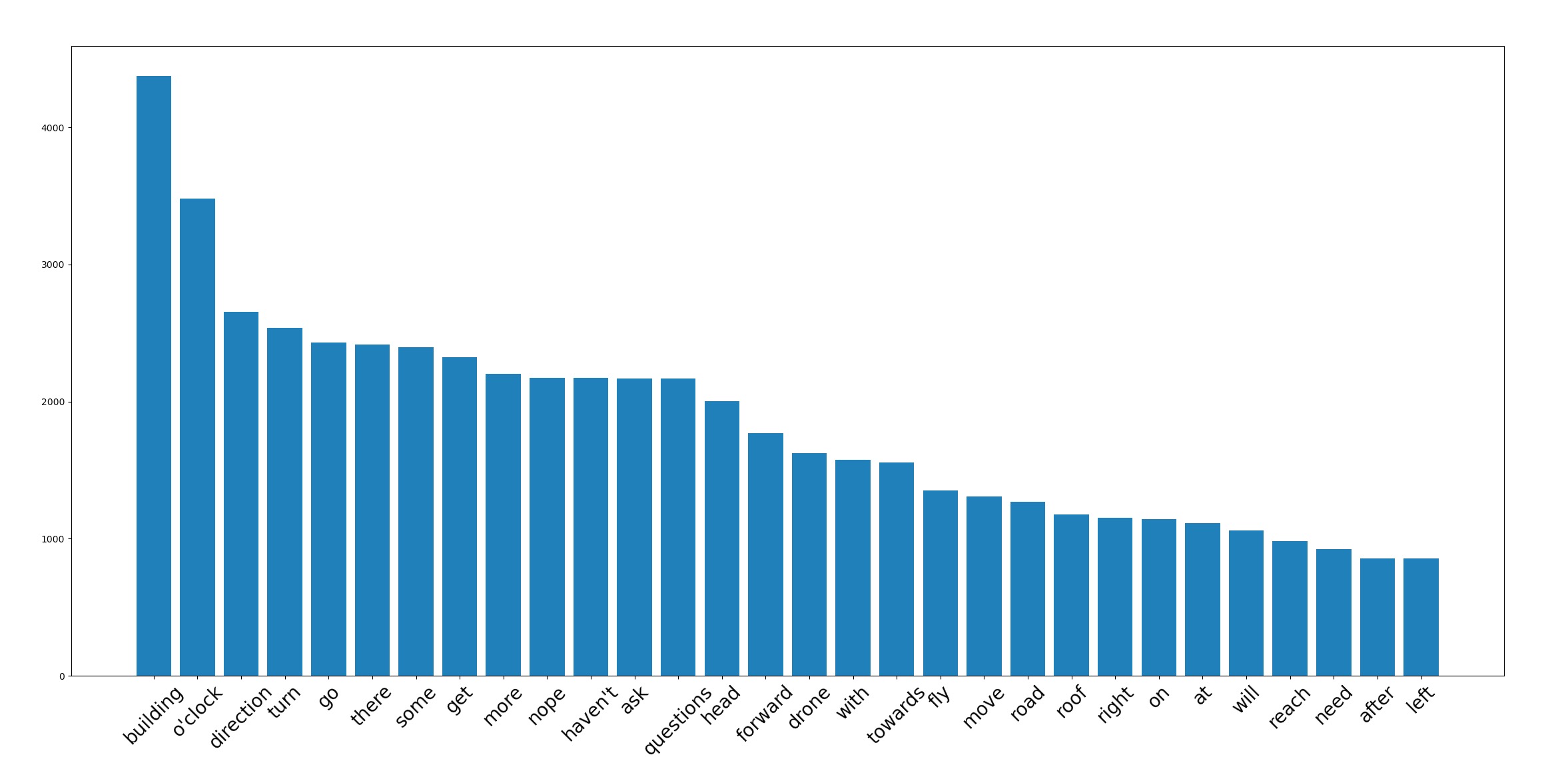}}
    
    \subfloat[Words from follower utterances]{\includegraphics[width=1\linewidth]{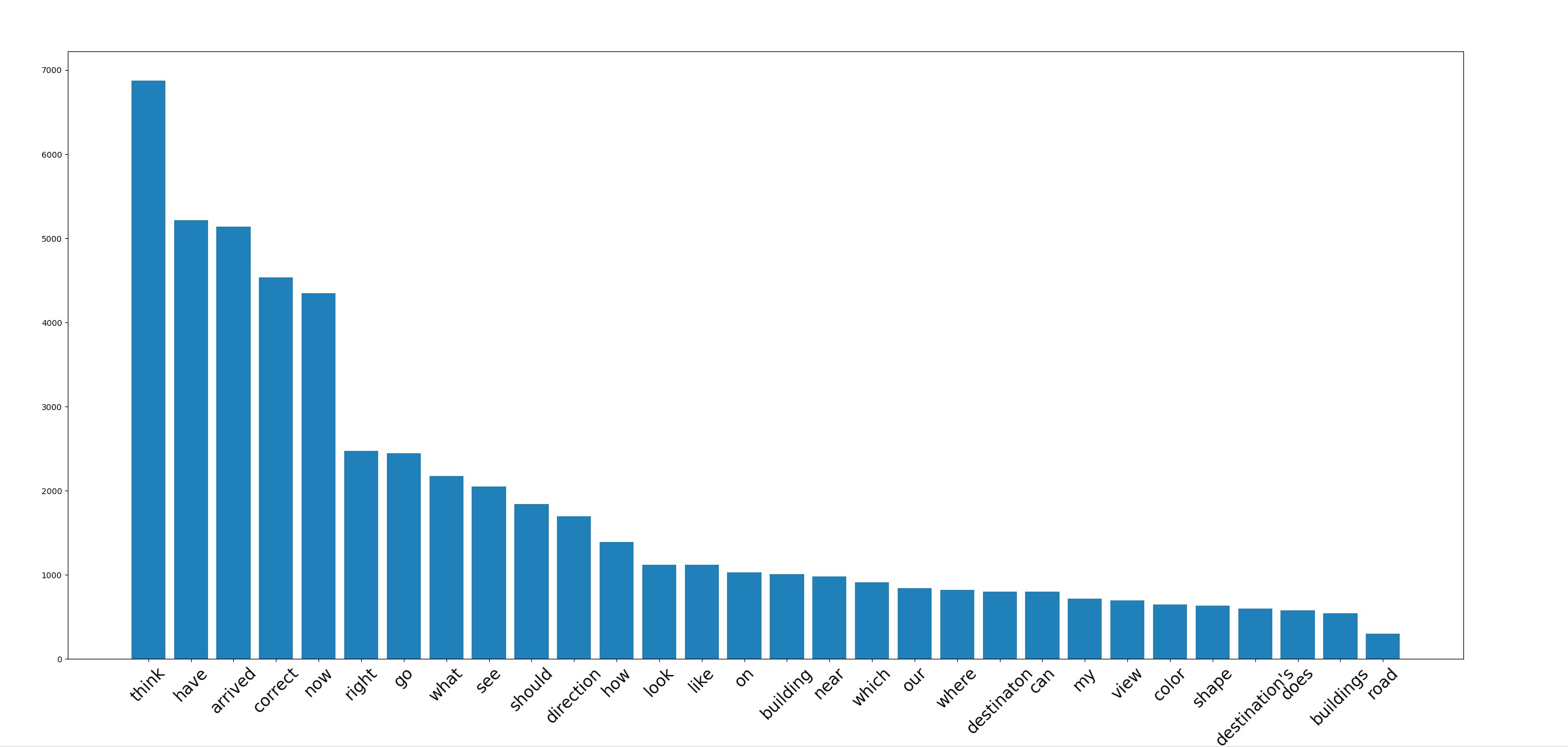}}
    
    \caption{Counts of top 50 most frequently used words in commander and follower utterances.}
    \label{fig:word_count}
\end{figure}

\subsection{Dialog Quality}
To ensure the dialogs in our dataset have good quality, we make efforts during the data collection process and conduct extra examination for the dialog data after the data collection.

During the data collection, online workers from Amazon Mechanical Turk (AMT) are playing as commanders and provide instructions in the dialog, who, compared with the follower that we hired to work on-site and supervised by us in-person, have a higher chance of generating low quality and incorrect language instructions. We develop some strategies to deal with these undesired instructions. First, if the follower, guided by the instruction, lets the drone navigate to a direction that is more than 90 degrees different from the ground truth direction of the destination area, our simulator will automatically label the instruction as incorrect. Those labeled instructions will be discarded and collected again. Then, since the follower needs to read and understand the instructions, they have the chance to report the instructions as being low-quality or incomprehensible and skip them. 
Finally, in the remaining instructions that are not spotted as low-quality or incorrect, it is still possible that instructions are not accurate or incorrect due to human mistakes from the AMT workers, such as in Figure \ref{fig:app_example_3}. By manually checking the dialogs and navigation trajectories in randomly selected subsets of our AVDN dataset, we spot only 5 instructions with potential mistakes in 50 dialogs. In those cases, because the follower successfully followed the instruction, we keep those instructions unchanged even if they didn't help guide the follower to find the destination area. In the real world, the user in AVDN could also make mistakes, so this mistake tolerance strategy makes our dataset even closer to real scenarios.

We further examine the dialog quality after the data collection by analyzing the dialogs. The average utterance (human-written instructions and questions) in a dialog is 3.1, with a minimum and maximum being 1 and 7 because each dialog includes at least one instruction written by a human. The average number of words written by commander and follower are 45 and 19, and there are about 15 words from auto-instructions. Also, in Figure \ref{fig:word_count}, we show the distribution of the top 30 most frequent words in the commander's and follower's utterances. The results show a smooth variance across nouns, verbs, adjectives, and prepositions, indicating that our dataset's utterances have rich contents and good variety. Last but not least, we manually checked the dialogs in all validation and test sets by visualizing the corresponding navigation trajectory and the dialog, and we observed no major issue.

\section{Interface for workers in dataset collection}
We use help from Amazon Mechanical Turk (AMT) workers and human drone experts during the collection of our Aerial Vision-and-Language Navigation (AVDN) dataset, where the AMT workers play the commander role providing instructions the drone experts play the follower role asking questions and controlling the drone. In this section, we demonstrate the interface for both groups of workers with all the information they receive in the data collection procedure. 

\subsection{Interfaces for commanders}
\begin{figure*}[t]
    \centering
    \includegraphics[width = 0.75\linewidth]{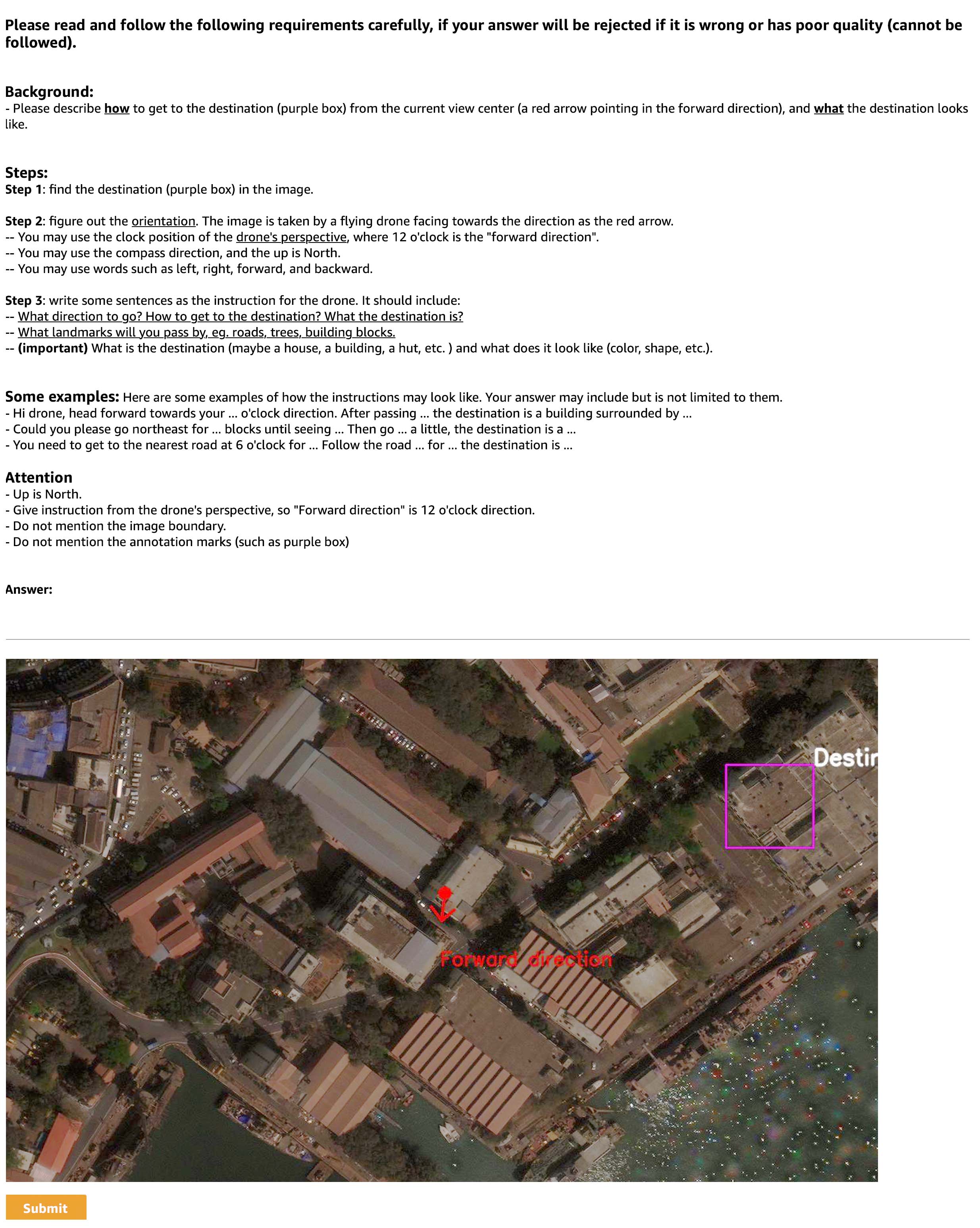}
    \caption{Interface for AMT workers (commanders) in first round of data collection.}
    \label{fig:commander_1}
\end{figure*}

\begin{figure*}[t]
    \centering
    \includegraphics[width = 0.7\linewidth]{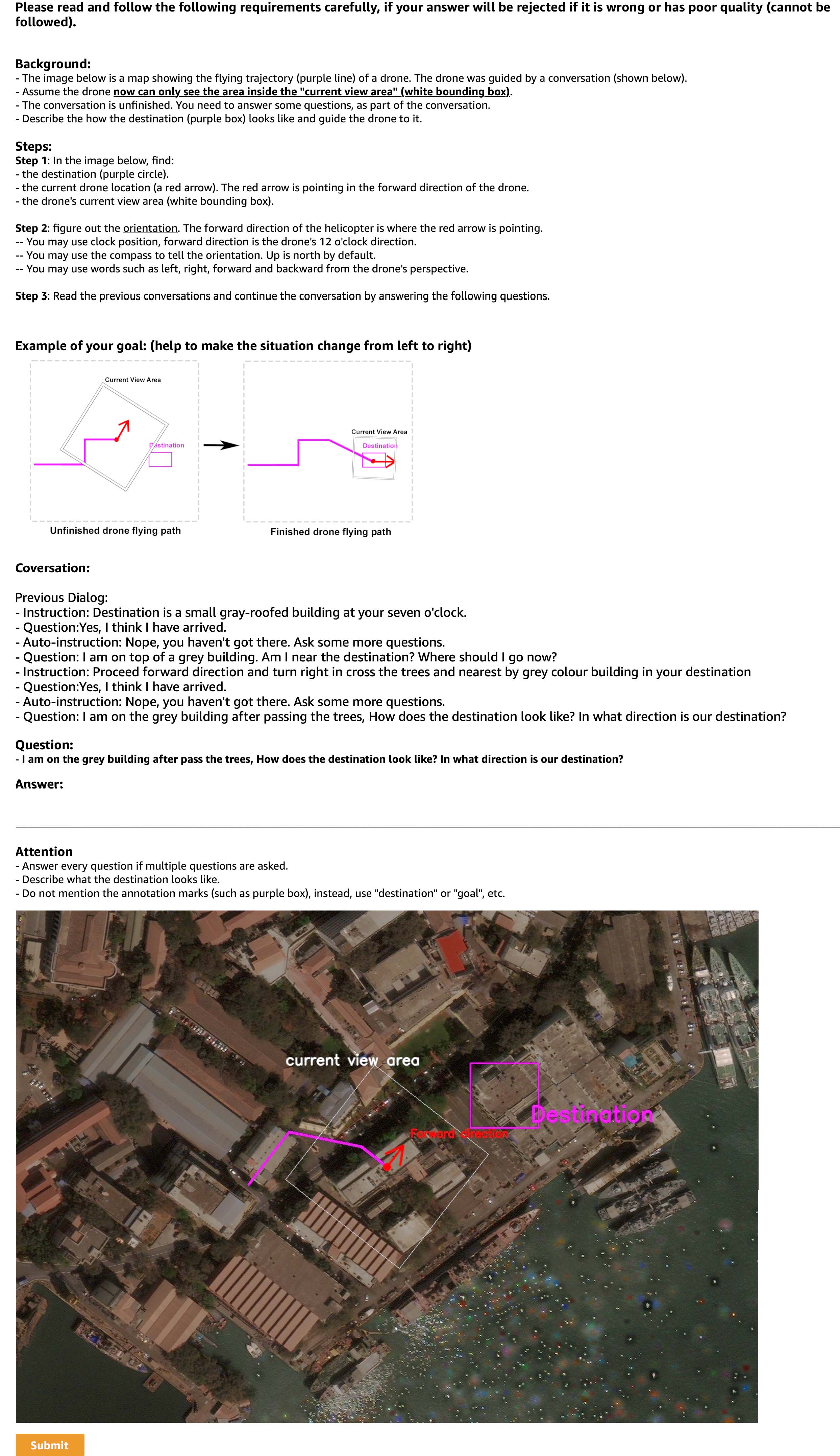}
    \caption{Interface for AMT workers (commanders) in following rounds of data collection.}
    \label{fig:commander_2}
\end{figure*}
There are two interfaces for commanders (AMT workers) depending on which data collection round it is. The interface includes one trajectory each time and contains all the information needed for the commander to create the instruction. Detailed and step-by-step instructions for what needs to be done as a commander are introduced at the beginning of the interface. The AMT workers need to write sentences in the {\it Answer} according to the provided information.

In the first round of data collection, the commander needs to write the initial instruction based on an overview of the AVDN trajectory. As shown in Fig. \ref{fig:commander_1} the satellite image shows the trajectory overview marked with a predefined staring position (the red point with an arrow showing the drone's direction at the starting position) and a destination area (purple bounding box).

In the data collection round after the first round, the commander is required to give follow-up instructions, i.e., answers, to the questions from the follower. The user interface for the second and following rounds is shown in Fig. \ref{fig:commander_2}. Besides all the information shown to the commander in the first round, the follower is also provided with previous dialog, past trajectories (broken purple line), and the view area corresponding to the most recent time step (named current view area marked with white bounding box).

\subsection{Interface for followers}

\begin{figure*}[t]
    \centering
    \includegraphics[width = 0.7\linewidth]{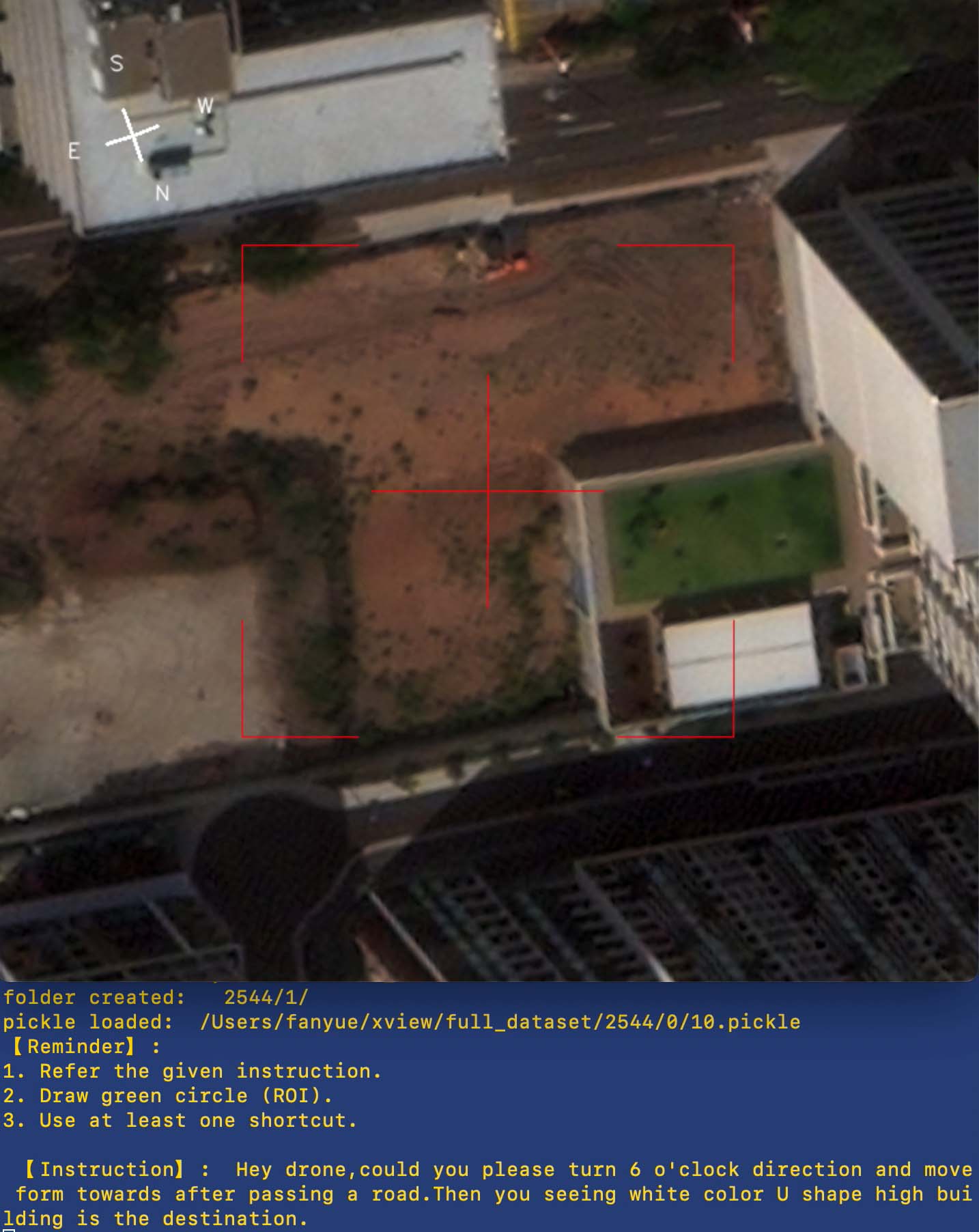}
    \caption{Interface for human drone experts (follower). The upper window shows the simulated drone's visual observation and the lower window shows the previous dialog. }
    \label{fig:follower_0}
\end{figure*}

The follower uses an interface to interact with our simulator. In our simulator, they receive instructions from the commander and control the simulated drone. The keyboard is used to simulate the drone controller with eight keys representing four channels in the controller, where key {\it w} and {\it s} represent the channel controlling forward and backward movement, key {\it a} and {\it d} represent the channel controlling left and right movement, key {\it q} and {\it e} represent the channel controlling rotating clockwise and anti-clockwise movement and key {\it 1} and {\it 2} represent the channel controlling altitude change. After the experts finish the control, the commander can either claim the destination is reached or ask questions for more instruction. As in Fig. \ref{fig:follower_0}, the interface is an image window showing the simulated drone's visual observation and a text window for displaying the previous dialogs and inputting questions from the follower. There is a compass on the top left of the image window, showing the orientation of the simulated drone. The red cross in the image window shows the center of the view, helping the follower control the drone to right above the destination area, and the red corners in the window show the area of 0.4 IoU with the view area. The follower is instructed to make the destination area larger than the area indicated by the red corners in order to finish successful navigation.

\end{document}